\documentclass[sn-mathphys]{sn-jnl}

\jyear{2021}%
\theoremstyle{thmstyleone}%
\usepackage{cleveref}
\theoremstyle{thmstyletwo}%

\theoremstyle{thmstylethree}%

\begin{document}

\title[Semi-Supervised Visual Tracking of Marine Animals]{Semi-Supervised Visual Tracking of Marine Animals using Autonomous Underwater Vehicles}


\author*[1]{\fnm{Levi} \sur{Cai}}\email{cail@mit.edu}

\author[2]{\fnm{Nathan E.} \sur{McGuire}}\email{nmcguire@whoi.edu}

\author[3]{\fnm{Roger} \sur{Hanlon}}\email{rhanlon@mbl.edu}

\author[4]{\fnm{T. Aran} \sur{Mooney}}\email{amooney@whoi.edu}

\author[2]{\fnm{Yogesh} \sur{Girdhar}}\email{ygirdhar@whoi.edu}

\affil*[1]{\orgdiv{Massachusetts Institute of Technology and Woods Hole Oceanographic Institution Joint Program}, \orgaddress{\city{Woods Hole}, \postcode{02543}, \state{MA}, \country{USA}}}


\affil[3]{\orgname{Marine Biological Laboratory}, \orgaddress{\city{Woods Hole}, \postcode{02543}, \state{MA}, \country{USA}}}

\affil[4]{\orgdiv{Biology Dept.}, \orgname{Woods Hole Oceanographic Institution}, \orgaddress{\city{Woods Hole}, \postcode{02543}, \state{MA}, \country{USA}}}

\affil[2]{\orgdiv{Applied Ocean Physics and Engineering Dept.}, \orgname{Woods Hole Oceanographic Institution}, \orgaddress{\city{Woods Hole}, \postcode{02543}, \state{MA}, \country{USA}}}


\abstract{In-situ visual observations of marine organisms is crucial to developing behavioural understandings and their relations to their surrounding ecosystem. Typically, these observations are collected via divers, tags, and remotely-operated or human-piloted vehicles. Recently, however, autonomous underwater vehicles equipped with cameras and embedded computers with GPU capabilities are being developed for a variety of applications, and in particular, can be used to supplement these existing data collection mechanisms where human operation or tags are more difficult. Existing approaches have focused on using fully-supervised tracking methods, but labelled data for many underwater species are severely lacking. Semi-supervised trackers may offer alternative tracking solutions because they require less data than fully-supervised counterparts. However, because there are not existing realistic underwater tracking datasets, the performance of semi-supervised tracking algorithms in the marine domain is not well understood. To better evaluate their performance and utility, in this paper we provide (1) a novel dataset specific to marine animals located at \url{http://warp.whoi.edu/vmat/}, (2) an evaluation of state-of-the-art semi-supervised algorithms in the context of underwater animal tracking, and (3) an evaluation of real-world performance through demonstrations using a semi-supervised algorithm on-board an autonomous underwater vehicle to track marine animals in the wild.}




\maketitle

\section{Introduction}\label{sec1}

\begin{figure}[ht!]
    \centering
    \includegraphics[width=\textwidth]{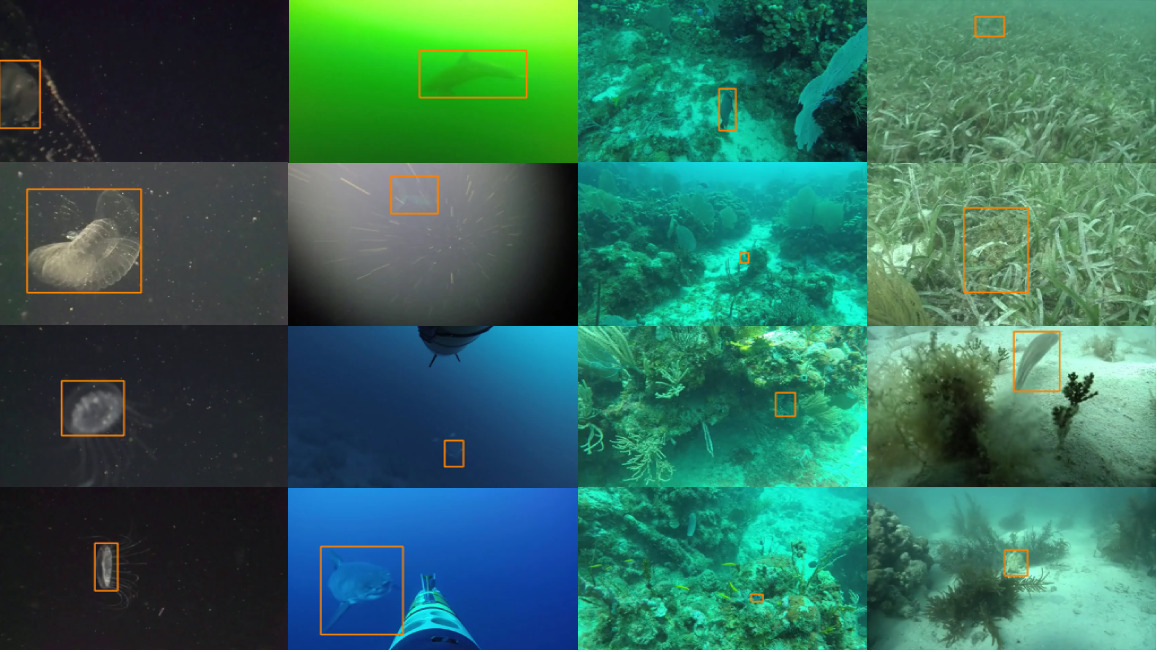}
    \caption{We present an initial dataset for evaluating performance of visual semi-supervised trackers for tracking marine animals in the wild. This dataset attempts to capture difficulties (and sometimes benefits) in tracking across species, environments, and behaviors. It consists of 33 densely-labelled sequences, collected by both divers and AUVs in real marine animal tracking scenarios, averaging over 1-minute in length.}
    \label{fig:dataset_overview}
\end{figure}

This work proposes the use of \textit{semi-supervised visual tracking} (SST) algorithms on \textit{autonomous underwater vehicles (AUV)} to track marine animals with no prior training. Semi-supervised algorithms have shown remarkable success in generic tracking benchmarks, but these benchmarks do not provide sufficient evidence of their performance in the underwater domain. This work aims to first establish the effectiveness of semi-supervised trackers in marine tracking tasks through domain-specific benchmarking, and in addition, demonstrate the use of a semi-supervised tracker in the real-world. Our contributions are thus to provide (1) a unique and underwater-specific dataset consisting of videos of mobile marine animals in their natural environment taken by following them with a moving camera system, (2) an evaluation of current state of the art semi-supervised tracking algorithms on this dataset using metrics relevant to the problem of marine animal tracking, and (3) a novel robotic demonstration of using a semi-supervised tracker in the real-world by deploying it on an AUV to track a marine animal in the wild. 

\textbf{Visual tracking of marine animals} In-situ visual observations of marine organisms can provide valuable insights into their biology that can be difficult to discern using other modes of observations. These observations can be used to characterize details of animal behavior and their interactions within that ecosystem. However, gathering in-situ observations using current approaches, especially in the case of marine animals, can be expensive, time-restrictive due to depth effects, and often dangerous. To achieve these observations, marine biologists have long relied on diver-based operations \cite{hanlon_crypsis_1999, hanlon_flamboyant_2020}, tags \cite{kukulya_3d_2015, mooney_biologging_nodate}, and occasionally using human-occupied vehicles (HOVs) \cite{priede_abyssal_2020} or remotely operated vehicles (ROVs) \cite{katija_visual_nodate-1} to gather visual observations. Each of these approaches is uniquely suited depending on the animal, environment, costs, hazards, and equipment that are present. 

Quantifying behavior requires long sequences of behavioral interactions and this becomes increasingly difficult with mobile animals. One of the most productive methods of measuring behavior is known as "focal animal sampling" in which video is acquired in a very disciplined manner by continually filming either a single animal, or pairs of animals for long periods to enable quantification and statistical analyses \cite{bateson_2021}. The reason for this is that key behaviors are not predictable and thus large video data sets over long continuous periods are required to capture both ongoing and episodic events. Divers become exhausted and cold with this demanding method, and the animal tracking can lead them to deeper water or beyond safe retreat to the surface vessel. Moreover, the bubbles from SCUBA and the changing shape of the diver can bias the target species' behavior, thus negatively influencing the natural interactions of the target species. Manually controlled ROVs can assist to a degree but they are prone to being pulled by currents that push the tether and visually lose track of the target species. Moreover, ROVs often cannot be deployed close to the seafloor in complex coral reef like environments which can pose significant danger to both the robot (entanglement) and the reef.  ROVs have been adapted to record static benthic animals with a regimented sampling routine \cite{williams_surveying_2009} but the ability to follow mobile animals is far more challenging. AUVs have the potential to follow and record mobile animals without the tether problem.

Autonomous underwater vehicles (AUVs) equipped with cameras and higher powered computers have potential to greatly supplement visual observations of animal behaviors in more difficult to access regions of the ocean, or to provide longer term observations in cases with limited human support. However, fully autonomous AUVs can currently only track a limited set of organisms in a small set of environments. Currently, trackable animals (i) are either tagged \cite{kukulya_3d_2015}, meaning a device must be physically attached to them that emits an acoustic signal to localize against, (ii) are found in relatively simple visual situations (such as the deep midwater column)\cite{yoerger_hybrid_2021}, or (iii) have significant amounts of labelled visual data available \cite{katija_visual_nodate-1}. These can be limiting in many circumstances because tags are difficult to install, most marine animals live in visually complex environments ranging from coral reefs to hydrothermal vents, and the underwater community generally has limited labelled data available for many species of interest.

\textbf{Semi-supervised trackers} Semi-supervised trackers have been proposed as alternatives to fully-supervised tracking methods because they require \textit{no target-specific pretraining}, are able to \textit{run in real-time}, and have shown high accuracy on a number of difficult visual object tracking benchmarks. Semi-supervised tracking algorithms, at run-time, only require an initial bounding box of the target to be provided by a human, and then the algorithm autonomously localizes the target in all subsequent frames. Their performance on existing benchmarks suggests they may be useful to the AUV community, by providing an alternative strategy to enable mostly autonomous visual data gathering without the need of significant visual training data or additional tracking equipment. This means AUVs could track a more diverse set of animals and in a wider range of environments. However, these methods have only been evaluated on generic datasets, which have few realistic examples of tracking animals fully underwater, so their effectiveness in these environments is not well known. 

\textbf{Challenges unique to the underwater domain} Underwater environments have many unique visual characteristics that are under-represented in generic datasets that are used to evaluate semi-supervised trackers. Some examples include: depth and distance dependent color absorption, presence of marine snow, extreme deformations and body shapes of marine animals, camouflaging, and large variations in visual complexity of underwater habitats forming the background. Hence a dedicated evaluation of their effectiveness specifically in realistic underwater settings is necessary. For instance, while the largest existing generic dataset, LaSOT \cite{fan_lasot_2020}, contains many examples of underwater animals, these sequences are primarily taken above-water or in aquariums with idealized lighting. Although there are a few underwater tracking sequences, they contain mostly turtles in relatively clear and simple cases. 

The most immediate visual distinctions are caused by water as a medium itself, which causes light distortion and absorption. In the former, following Snell's law, light traveling between media of varying densities causes light to bend, this is prevalent in cases where a camera is placed inside a waterproof housing, and so light must pass from water, to glass or plastic, and then air before reaching the camera lens. In addition, water absorbs different frequencies of light at different rates \cite{akkaynak_sea-thru_nodate}, red light being the most easily absorbed and blue light the last to be absorbed. This causes significant loss of color-based information in environmentally-lit situations, especially deeper underwater. Underwater caustics near the surface, where light refracts and reflects underwater causing bright and constantly moving patterns, can also serve as significant distractors.

In deeper waters or at night, where environmental light is less of a concern, active lighting from a diver or vehicle can simplify the color absorption problem. Unfortunately, such relatively bright lights override the dark adaptation of the target animals night vision, and thus affect its behavior. In addition, again due to light absorption, visible distances are typically short. Furthermore, marine snow, small biological particulate matter, is prevalent in the ocean \cite{wang_underwater_2021}, and active light reflects off of these particles. This can cause either general haziness or become sharp and ubiquitous visual features.

The types of habitats and the animals themselves are also unique and present significant challenges. Habitats such as coral reefs, mid-water, hydrothermal vents, kelp forests, sea grass, etc. provide various visual challenges, several of which are shown in \Cref{sec:dataset}. Marine organisms exhibit a variety of different swimming, foraging and defensive behaviors that can disrupt the capabilities of imaging to follow them. Camouflage by many marine animals is highly advanced and diverse. An extreme example among invertebrates are octopuses, cuttlefish and squid which can instantly change their camouflage pattern, and even their body shape as well as their skin 3D texture, all of which is exceptionally difficult to discern with digital imagery. These animals as well as many fishes have specific behaviors to actively avoid detection and tracking, through camouflage, inking, darting, hiding, and so forth. 

Finally, collection of raw data can be especially difficult underwater, and so too is the development of large databases of imagery. Although extensive datasets exist from the underwater domain, they are either aiming a stationary camera to a stationary subject, or aiming at characterizing fish biodiversity with wide angle lenses and thus not focusing on single target tracking.  

Depending on depth and distance from shore or the hazards of the environment or animals themselves, significant equipment, extensive training, and labor could be required. Even in relatively shallow water, below 10m, longer-term tracking of marine animals requires a SCUBA certification and specific camera housings. If the animal varies its depth quickly, this can be dangerous to the diver. In other instances, only rare vehicles are capable of collecting data, for instance in the deep sea.

In this paper, we review related works in \Cref{sec:related_work}, present the dataset and evaluation results in \Cref{sec:dataset}, present the real-world AUV tracking results in \Cref{sec:stingray}, and discuss the results and their implications in \Cref{sec:discussion} and finally give concluding remarks in \Cref{sec:conclusion}.


\section{Related Work}
\label{sec:related_work}

This work is most closely related to works in autonomous vision-based tracking for marine animals and the development of datasets for evaluating real-time, semi-supervised tracking methods. In the following section we discuss how this work is situated in these contexts.

\subsection{Autonomous vision-based marine animal tracking}

We focus on the current use of AUVs for marine animal tracking. We first discuss \textit{passive} video tracking of animals. In passive contexts, video cameras collect data of marine animals, but the \textit{images themselves} are not used to inform AUVs where to look. These strategies tend to fall in two categories: surveys and acoustic tag-based tracking. In the former, vehicles such as the MBARI i2MAP Dorado or the Seabed AUV \cite{williams_surveying_2009} perform pre-programmed surveys in ocean, collecting video along the pre-determined track, which is analysed afterwards. Other passive visual observation gathering by AUVs is accomplished through acoustic tags. Animals such as sharks, cetaceans, penguins, turtles, some larger fish, etc. can be outfitted with an acoustic tag that an AUV can then use to localize its position relative to the vehicle. These AUVs, such as the REMUS series, are then equipped with cameras, often many oriented in several directions, that record video to a memory card and again analysed after the mission \cite{kukulya_multi-vehicle_2016}. Tags are only able to be attached to specific species of animals, and typically require significant deployment effort and training to affix. In addition, videos collected in this manner have less quality guarantee, and animals may not stay in the frame of a single camera for very long. 

AUVs equipped with \textit{active} visual tracking capabilities have recently been developed. The Mesobot platform \cite{yoerger_hybrid_2021} is developed to track slow-moving animals in the mesopalegic zone (300m-1000m depths), also known as the Ocean Twilight Zone, which has very little light. This system consists of a grey-scale, stereo-camera system for tracking and provides its own light. Animals are tracked through established color segmentation and blob-tracking methods, and the Mesobot was used to successfully track jellyfish and larvaceans in-situ for several hours. However, these methods only work in very simple tracking scenarios, which is unique to the mesopalegic zone and the types of animals that live there, where jellyfish illuminated by on-board lights are easily distinguished from the dark ocean backdrop. Katija et al. \cite{katija_visual_nodate-1} introduced the use of tracking-by-detection and deep learning strategies to increase robustness and track in more complicated scenarios and demonstrated its effectiveness on-board the MBARI MiniROV. In this case, a deep convolutional neural network, RetinaNet \cite{lin_focal_nodate} is pre-trained on the target(s) of interest. During run-time, the network is used to detect targets, and a data association strategy is used to determine which detections correspond to the appropriate target. In both of these scenarios, once the target is localized in each frame, the vehicle is commanded to update its position to center the target in the camera frame.

These systems have provided invaluable insights for researchers, however, Mesobot is only able to track simple organisms in the deep sea and the MBARI MiniROV can only track animals for which they have significant training data already collected. Our work aims to show if semi-supervised trackers can enable AUVs to track a much larger range of animals and in more varied habitats where significant target-specific training data for fully supervised neural networks is not available.

\subsection{Underwater visual datasets for marine animal tracking}

Due to the popularity of machine learning methods for land-based animal classification, several datasets have recently been developed for marine animal classification as well. Most prominent among these are FathomNet, VIAME, AIMs Ozfish, MBARI VARS, and DeepFish \cite{katija_fathomnet_2022, dawkins_open-source_2017, schlining_mbaris_2006, noauthor_ozfish_nodate-1, saleh_realistic_2020-1}. All of these datasets are useful for training deep learning-based networks for classification and can be incorporated into fully-supervised trackers. However, each is highly localized to a specific set of animals. For instance, AIMs Ozfish and DeepFish both are collected from nearby Australia, and primarily contain images of vertebrate fish. VIAME is a more general underwater animal dataset, but mostly contains imagery from smaller organisms that visit baited camera traps. In addition, these types of sequences are insufficient for evaluating tracking scenarios with moving cameras, or for organisms in the open ocean. FathomNet and the MBARI VARS datasets are perhaps the only datasets that explicitly attempt to capture data useful for evaluating active tracking tasks. However, the MBARI VARS dataset is difficult to access publicly, and FathomNet's dataset, at the time of this writing, only includes deep ocean species such as jellyfish, larvaceans, and slow-moving seafloor organisms found off the coast of Monterey Bay in California. 

None of these datasets is large enough to provide robust training for fully-supervised tracking methods to work on the full range of marine animals of interest, such as those illustrated in \Cref{tab:seq-list}, though they are invaluable starts. Furthermore, none provide longer video sequences that allows evaluation of semi-supervised trackers on realistic tracking tasks where there is significant camera motion, high-frame rates of at least 10fps, which are necessary to enable reasonable feedback control in AUVs, and a realistic set of behaviors where baited traps are not used.

\subsection{Deep learning approaches for semi-supervised tracking}

While semi-supervised visual tracking has a fairly long literature, only recently has it started to gain more widespread attention with the introduction of deep learning-based feature extractors and classifiers. One of the first deep learning-based SSTs, MDNet \cite{nam_learning_2016}, achieved the highest performance on an early generic object tracking benchmark VOT \cite{kristan_visual_2013}. However, MDNet could not run in real-time. Since then, several innovations, especially in deep learning-based trackers, have enabled SSTs to achieve real-time tracking speeds while continuing to consistently be top performers in accuracy on multiple generic object tracking benchmarks.

Of these innovations, two architectures, which we will refer to as either \textit{Siamese-based} and \textit{online discriminator-based}, in particular have dominated the most recent benchmarks. Tao et al. \cite{tao_siamese_2016} first introduced Siamese-based networks as a way to achieve the accuracy of deep learning-based trackers with real-time speed. In these architectures, a pre-trained deep neural network backbone is used to extract features from an initial template image of the target. Then using the backbone network, features are extracted from subsequent images, and the region (in the extracted feature space) that is most similar to that of the features of the template image, is labelled as the target of interest in that frame. Subsequent Siamese-based trackers innovated on the types of backbone networks used for feature extraction or downstream optimization tasks such as providing full masks or selecting between distractors \cite{li_siamrpn_2019, wang_fast_2018, li_siamrpn_2019}.

Siamese-based networks suffer in performance however during object appearance changes, since they rely only on the appearance of the object in the first frame for reference. Later, Danelljan et al. \cite{danelljan_atom_2019} introduced an online learning component in ATOM, where recent frames, in addition to the first frame, are used to train a discriminator that is then used to classify subsections of the current tracking frame as the target or not. This further improved performance in the case of appearance changes, and many later trackers added post-processing steps to gain additional performance boosts \cite{bhat_learning_2019, danelljan_convolutional_2015, danelljan_eco_2017, chen_transformer_2021, wang_transformer_2021}. 

It is interesting to note, as described in \cite{wang_transformer_2021}, that Siamese-based networks are effectively special cases of online discriminator-based architectures, where the online learning step is removed, which may be important in how they are analysed. In particular, Siamese-based networks are generally robust in the long-term, meaning if the target appearance returns to something similar to the first frame, the tracker can recover. In contrast, in the online discriminator case, because later frames are not labelled, small errors in appearance-learning accumulate, and can cause the tracker to drift from focusing on the correct features resulting in lower long-term robustness \cite{mueller_benchmark_2016-1}. However, when this tradeoff is carefully addressed, the online visual tracking problem seems to tend towards online learning approaches, because the best predictor of an object's appearance at time $t$ is its appearance at $t-1$.

In some of these cases, trackers can catastrophically fail either from appearance change or from self-drift, even when the object remains in the sequence at all times. We thus believe it is important for evaluation datasets to contain longer (in duration) sequences, where object appearances change over time, in order to verify these issues.

\subsection{Generic benchmarks for semi-supervised tracking}

Many general purpose semi-supervised tracking datasets and benchmarks exist, such as LaSOT, GOT10K, VOT2020, DAVIS, OxUvA, OTB100, TrackingNet, Youtube-BB, NfS, and UAV123 \cite{fan_lasot_2020,huang_got-10k_2021,kristan_eighth_2020, wu_object_2015, caelles_2019_2019,xu_youtube-vos_2018-1, mueller_benchmark_2016-1,galoogahi_need_2017,valmadre_long-term_2018}. However, as noted in LaSOT, many of these datasets such as OxUvA, TrackingNet, and Youtube-BB, are only labelled intermittently, so many frames are not guaranteed to have a human-checked ground truth. This is not well-suited for evaluating trackers that need to run on vehicles where high-frame rates are required. As in LaSOT, we thus focus on comparing against the following densely labelled datasets, also noted in \Cref{tab:dataset-comp}:

\textbf{OTB100 \cite{wu_object_2015}, VOT2020 \cite{kristan_eighth_2020}, Need for Speed (NfS) \cite{galoogahi_need_2017}} are smaller standard datasets, OTB and VOT are some of the earliest benchmarks to have been introduced, with VOT2020 being a slightly updated version of VOT, using a different bounding box methodology and swapping out a few sequences. NfS was developed for evaluating extremely fast frame-rate systems, with frames at 240fps. However, in all 3 of these datasets combined, only 2 sequences are of underwater animals at all, both in fairly ideal lighting conditions.

\textbf{GOT-10K \cite{huang_got-10k_2021}} was developed to have by far the most sequences of other current datasets, to be used for both training and evaluation purposes. Some of these sequences also include underwater animal tracking. However, these sequences are very short, on average only 14.9 seconds in duration across the whole dataset, or only 9.5 seconds in duration when considering animal classes. These are too short to effectively evaluate performance for longer-term deployments on real vehicles.

\textbf{UAV123 and UAV20L\cite{mueller_benchmark_2016-1}} are part of the same dataset, developed to evaluate tracking of humans and vehicles from on-board unmanned aerial vehicles (UAVs). UAV20L consists of 20 sequences that are longer in length than the rest of the UAV123 sequences, and is most like our dataset in terms of its overarching goals. However, this dataset is only focused on humans and ground vehicles.

\textbf{LaSOT \cite{fan_lasot_2020}} contains 1550 sequences that are densely labelled, at 30fps, and capture many realistic tracking scenarios, and is likely most useful for evaluations similar to ours. However, among those sequences, while many are taken of underwater situations, most occur in artificial environments such as aquariums, or are above the surface of the water. We found only 12 sequences that were consistent with our goals, but do not have species diversity, as 11 are of turtles and 1 of an alligator.

For instance, LaSOT focused on longer sequences, GOT10K increased number of classes, OxUvA provided significant numbers of fully occluded or fully lost targets, etc. Our approach most resembles UAV20L, which aims to provide a more domain-specific dataset with longer tracking sequences that are realistic to the overall tracking problem on a mobile platform. 

While LaSOT and UAV20L closely resemble our tracking evaluation goals, they do not have sufficient coverage of the underwater tracking domain to provide convincing evaluations of the range of underwater-specific issues.

\section{The Visual Marine Animal Tracking Dataset}\label{sec2}
\label{sec:dataset}

\begin{table}[]
\resizebox{\textwidth}{!}{%
\begin{tabular}{@{}lrrrrrrrr@{}}
\toprule
 &
  \multicolumn{1}{r}{OTB100} &
  \multicolumn{1}{r}{VOT2020ST} &
  \multicolumn{1}{r}{NfS} &
  \multicolumn{1}{r}{GOT-10K} &
  \multicolumn{1}{r}{LaSOT} &
  \multicolumn{1}{r}{UAV123} &
  \multicolumn{1}{r}{UAV20L} &
  \multicolumn{1}{r}{VMAT (Ours)} \\ \midrule
Num sequences     & 100   & 60   & 100   & 9695  & 1550  & 123   & 20    & 33    \\
Frame rate        & 30    & 30   & 240   & 10    & 30    & 30    & 30    & 30    \\
Min duration (s)  & 2.4   & 1.4  & 0.7   & 2.9   & 33.3  & 3.6   & 57.2  & 14.6  \\
Mean duration (s) & 19.7  & 11.1 & 16.0  & 14.9  & 83.4  & 30.5  & 97.8  & 74.9  \\
Max duration (s)  & 129.1 & 50.0 & 86.1  & 141.8 & 379.9 & 102.8 & 184.2 & 248.2 \\
Fully underwater animal sequences   & No    & Only 1   & Only 1    & Yes    & Only 1 species    & No   & No   & Yes   \\
Realistic tracking sequences     & Yes    & Yes   & Yes & No    & Yes    & Yes   & Yes   & Yes   \\ \bottomrule
\end{tabular}%
}
\caption{Comparison of densely-labelled and high-fps benchmark datasets for evaluating semi-supervised trackers. Here we consider fully underwater animal sequences where both the camera and animal are underwater in nature, and not in an aquarium or similarly artificial setting. Here we consider realistic tracking sequences those that are aimed at longer duration, densely-labelled, and $>$10 fps (for vehicle control purposes) sequences.}
\label{tab:dataset-comp}
\end{table}

\subsection{Design principles}

In order to evaluate expected performance of semi-supervised trackers in a variety of challenging marine tracking tasks, we needed to develop a new evaluation dataset. Our dataset, which we refer to as the Visual Marine Animal Tracking (VMAT) dataset, is used for evaluation only, and is domain-specific, similar to the UAV20L dataset \cite{mueller_benchmark_2016-1} in terms of size and scope, but targeted at AUV deployments for tracking underwater animal targets. We thus focus on collecting data that is diverse across animals, habitats, behaviors, and tracking scenarios as possible, which we show in \Cref{tab:seq-list}. We also aim for each sequence to be as \textit{realistic} as possible, where a camera is actively tracking an animal, sequences are densely-labelled, where every frame is labelled, and are in as long duration as possible (without loss of sight). We ensure frame rates are over 10fps which we believe is the minimum frame rate necessary to perform reliable feedback control of most underwater vehicles, though higher is better.

\subsection{Data collection, processing, and annotation}

\begin{table}[]
\resizebox{\textwidth}{!}{%
\begin{tabular}{@{}rlllr@{}}
\toprule
\multicolumn{1}{l}{ID} & Animal & Habitat & Motion/Behavior & \multicolumn{1}{l}{\# Frames} \\ \midrule
0  & Octopus      & Seabed, sand \& grass  & Stop and go, fast & 1340 \\
1  & Octopus      & Seabed, sand \& grass  & Slow crawl        & 1129 \\
2  & Octopus      & Seabed, dense seagrass & Slow crawl        & 2971 \\
3  & Octopus      & Seabed, dense seagrass & Stop and go, slow & 2042 \\
4  & Shark        & Midwater, shallow, clear        & Constant swim     & 438  \\
5  & Shark        & Seabed, near-bottom    & Constant swim     & 1092 \\
6  & Shark        & Midwater, deep, marine snow  & Fast swim         & 750  \\
7  & Dolphin      & Midwater, turbid       & Fast swim         & 690  \\
8  & Larvacean    & Midwater, deep, marine snow  & Slow swim         & 601  \\
9  & Larvacean    & Midwater, deep, marine snow  & Slow swim         & 901  \\
10 & Jellyfish    & Midwater, deep, marine snow  & Slow swim         & 512  \\
11 & Jellyfish    & Midwater, deep, marine snow  & Slow swim         & 511  \\
12 & Striped Fish & Coral                  & Fast darting      & 750  \\
13 & Parrotfish   & Coral                  & Medium swim       & 2610 \\
14 & Parrotfish   & Coral                  & Medium swim       & 1080 \\
15 & Parrotfish   & Coral                  & Medium swim       & 1680 \\
16 & Lionfish     & Coral                  & Stationary        & 4770 \\
17 & Angelfish    & Coral                  & Medium swim       & 3870 \\
18 & Boxfish      & Coral                  & Medium swim       & 2580 \\
19 & Blue Tang    & Coral                  & Fast swim         & 1350 \\
20 & Blue Tang    & Coral                  & Fast swim         & 600  \\
21 & Squid        & Rocky Seabed           & Medium swim       & 2550 \\
22 & Squid        & Rocky Seabed           & Medium swim       & 5550 \\
23 & Octopus      & Rocky Seabed           & Crawling          & 2190  \\
24 & Snapper      & Coral                  & Fast swim       & 3180 \\
25 & Shark        & Coral                  & Medium swim       & 2490 \\
26 & Shark        & Coral                  & Medium swim       & 5514 \\
27 & Shark        & Coral                  & Medium swim        & 1566 \\
28 & Stingray     & Seagrass               & Medium swim       & 7448 \\
29 & Jack         & Coral                  & Medium swim       & 3960 \\
30 & Barracuda    & Coral                  & Fast swim         & 2363 \\
31 & Sea turtle   & Midwater               & Slow swim         & 3180  \\
32 & Jellyfish    & Seagrass               & Slow swim         & 1920 \\ \bottomrule
\end{tabular}%
}
\caption{List of sequences in the dataset and the corresponding sequence ID, animal, habitat, and behavior contained within the sequence.}
\label{tab:seq-list}
\end{table}

To build a diverse dataset, we gathered sequences from several different existing scientific sources, in addition to collecting our own sequences in the field. All sequences in the dataset were collected in active tracking scenario, where a single target is the focus for the entirety of the track. From each source, we selected the longest contiguous sequences that did not have full occlusions or loss-of-sight.

We began by collecting existing videos from nearby scientists. These consisted of difficult sequences of octopuses exhibiting a variety of behaviors (from highly conspicuous to effectively camouflaged) in seagrass, coral, and sandy habitats; these were filmed via SCUBA in Puerto Rico by co-author Roger Hanlon using a Panasonic HD HVX200. These sequences were obtained via a focal animal sampling routine for a study of adaptive camouflage.

To add midwater tracking scenarios with significantly larger and faster animals, Amy Kukulya and Roger Stokely, from the Woods Hole Oceanographic Institution, provided sequences of sharks and dolphins. These sequences were all collected by REMUS vehicles \cite{kukulya_3d_2015} tracking tagged animals. In these cases, even though the vehicles were actively tracking an acoustic beacon physically attached to the animals, they were only \textit{passively} collecting visual data. This meant that the animals rarely stayed in frame, so the sequences are relatively short. These sequences also show variation in lighting at different depths, as shallower tend to be blue, deeper are black, and water that is high in chlorophyll tends to be very green. It is also important to note the simpler backgrounds in these midwater examples. The deeper of these sequences also show significant marine snow. These can be seen in the second column of \Cref{fig:dataset_overview}.

We also then sought to add deeper ocean tracks. The Mesobot team, from their expeditions in the Pacific Ocean as published in \cite{yoerger_hybrid_2021}, were able to provide long tracks of midwater organisms in the ocean twilight zone (100m-3000m depths). These samples were collected fully autonomously by Mesobot via a standard blob tracking algorithm. While Mesobot sequences were longer in length, there do not include as much visual diversity, and so the included sequences are the ones that capture the most diverse motion instances.

Finally, to represent a larger range of species, the authors Levi Cai, Yogesh Girdhar, and Aran Mooney from the Woods Hole Oceanographic Institution, performed several SCUBA dive operations in the U.S. Virgin Islands, St. John in October 2021 and October 2022. We used GoPros in off-the-shelf underwater housings and followed a variety fish in coral reef, seagrass, and sandy environments. For each sequence, we track a single organism for as long as possible while minimizing occlusions. We found a new organism when it was no longer possible, and repeated this until our dive times were complete. Through these tracks, we collected 20 additional sequences across several species, including squid, an octopus, reef sharks, a stingray, a squid, and 9 types of reef fish. 

All videos were then reviewed manually by the author and the longest continuous sequences, without loss of sight and with continuing novel viewpoints or appearance changes, of each organism were selected. Because many tracker evaluation metrics are sensitive to differences in both spatial and temporal resolution we standardize all videos to a minimum common standard that is still usable for active AUV tracking. Specifically, we subsample frame rates down to 30fps and scale resolutions of all videos to 854x480 pixel (480p) using ffmpeg.

Finally, to produce dense ground-truth labels of each target, the authors manually labelled axis-aligned bounding boxes using LabelBox \cite{noauthor_labelbox_nodate} with manual labelling (adding a keyframe) or explicit review no less than every 15 frames. LabelBox then linearly interpolates between labels, and the authors do a last verification by watching the whole sequences. All sequences were labelled by Levi Cai and Yogesh Girdhar, and reviewed by Levi Cai.


\subsection{Dataset attributes and statistics for evaluation}

The overall dataset contains 33 sequences, with a total of 74K frames. On average, the sequences are 75 seconds long, with a longest sequence duration of 248 seconds. We emphasize the importance of longer \textit{duration} sequences to capture variability in animal behaviors. This is different from the \textit{NfS} philosophy which focuses on extremely fast behaviors. More general statistics and comparisons are shown in \Cref{tab:dataset-comp}.

After the selection process, the final dataset consists of 17 different types of marine animal: octopuses, sharks, dolphins, larvaceans, jellyfish, squid, turtle, stingray, and 9 species of fish (striped fish, parrotfish, lionfish, angelfish, boxfish, snapper, barracuda, jack, and blue tangs). They are distributed over several visually distinct marine habitats including coral reefs, sea grass, the shallow mid-water column, the deep mid-water column, and near sandy and rocky seabeds. And finally exhibit several types of swimming behaviors such as fast, medium, and slow constant swimming, darting, crawling, and stop and go maneuvers.

It is common to label sequences with attributes that may cause tracker difficulties and group evaluations by those attributes in order to determine which attributes cause issues during tracking. For this, we select 6 standard attributes including: scale variation (SV), low resolution (LR), partial occlusions (PO), difficult backgrounds (DB), and similar objects (SO), with a description of each and how they are determined in \Cref{tab:attributes}. Because of the inherent physical active tracking scenario and longer durations, all or most of the sequences in this dataset have camera motion, in-plane rotation, and show object appearance changes from rotation or deformation.

\begin{figure}[ht!]
    \centering
    \includegraphics[width=0.48\textwidth]{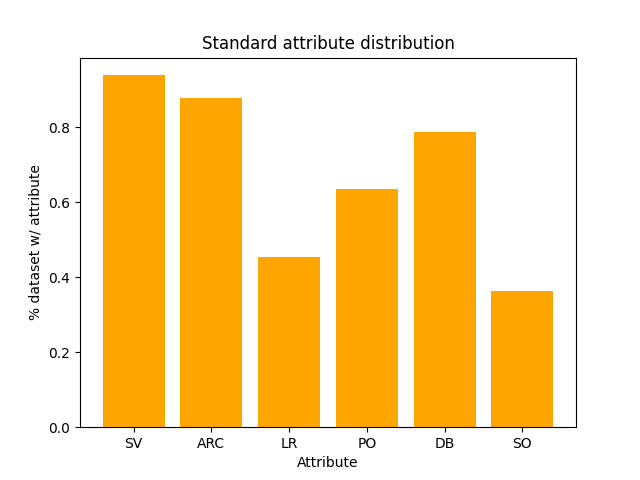}
    \includegraphics[width=0.48\textwidth]{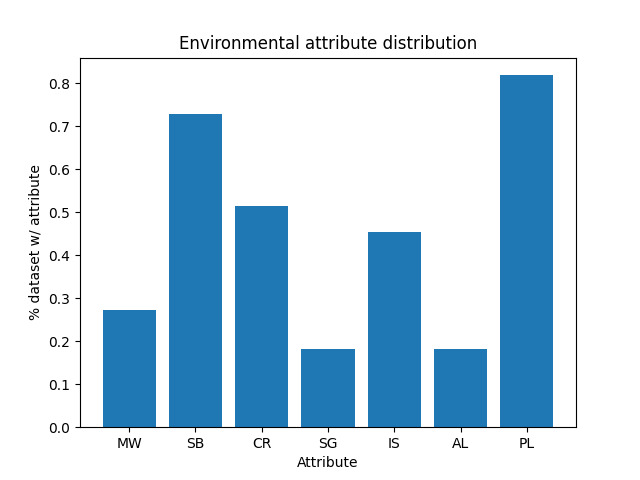}
    \caption{Distribution of sequences with each attribute in our dataset. On the left are standard generic attributes that we evaluate on and on the right are environmental-based attributes unique to our dataset.}
    \label{fig:attributes}
\end{figure}

In order to evaluate metrics that are more specific to the underwater domain and habitats, we include 7 additional attributes that were manually determined. These are midwater (MW), seabed (SB), coral reef (CR), seagrass (SG), intermittent sand or rocks (IS), and active lighting (AL) or passive lighting (PL). The descriptions of these attributes are also listed in \Cref{tab:attributes}. We note that CR, SG, and IS are all subsets of SB and are more notable for having DB. Because these environments are more typical in shallow regions, and because data was collected during the day, these also have PL. Likewise, MW environments tend to have simple backgrounds (they are generally blue or black), but may have more foreground clutter due to marine snow. In deeper MW environments, it is common to use AL, which amplifies the visual clutter due to marine snow, and so there is coupling inherent in these attributes that may be difficult to separate. A distribution of the sequences and associated attribute labellings is shown in \Cref{fig:attributes}.

\begin{table}[ht!]
\resizebox{\textwidth}{!}{%
\begin{tabular}{@{}lll@{}}
\toprule
Attr.      & Name                    & Description                                                    \\ \midrule
SV         & Scale Variation         & (Auto) Ratio of bbox px exceeds {[}0.5, 2{]} from initial bbox \\
ARC & Aspect Ratio Change & (Auto) Ratio of bbox aspect ratio exceeds {[}0.5, 2{]} from initial bbox \\
LR         & Low Resolution          & (Auto) Bbox is less than 1000px in area                                \\
PO         & Partial Occlusion       & (Manual) Object is partially occluded for more than 1 frame    \\
DB         & Difficult Background    & (Manual) Bbox overlaps with complex background                 \\
SO         & Similar Objects     & (Manual) Similar looking objects are nearby target object      \\ \midrule
Env. Attr. & Name                    & Description                                                    \\ \midrule
MW         & Midwater                & Target does not overlap with seafloor                            \\
SB         & Seabed                  & Target overlaps with seafloor at least half the sequence                                    \\
CR         & Coral Reef              & Target overlaps with coral reef at least half the sequence                           \\
SG         & Seagrass                & Target overlaps with seagrass at least half the sequence                             \\
IS         & Intermittent Sand/Rocks & Target overlaps with sand/rock at least half the sequence           \\
AL         & Active Lighting         & Scene is predominantly illuminated by AUV/diver                \\
PL         & Passive Lighting        & Scene is predominantly illuminated by environment (sun)        \\ \bottomrule
\end{tabular}%
}
\caption{Attributes selected for benchmarking. The first six attributes are common to all object tracking datasets and the latter seven are peculiar to the underwater environment. SV, ARC, and LR were automatically computed from characteristics of the labels of the corresponding video sequences. The remaining attributes were derived manually because they require some qualitative interpretation.}
\label{tab:attributes}
\end{table}

\section{Evaluation of state-of-the-art semi-supervised trackers}

\begin{table}[ht!]
\resizebox{\textwidth}{!}{%
\begin{tabular}{@{}lrrrr@{}}
\toprule
Tracker   & Abbrev. & \multicolumn{1}{l}{Source} & Arch. (S/OD) & Trans. layer? \\ \midrule
ECO \cite{danelljan_eco_2017}       & EC      & CVPR17                     & S            & N             \\
DaSiamRPN \cite{zhu_distractor-aware_2018} & DS      & ECCV18                     & S            & N             \\
ATOM \cite{danelljan_atom_2019}     & AT      & CVPR19                     & S            & N             \\
SiamRPN++ \cite{li_siamrpn_2019} & SR      & CVPR19                     & S            & N             \\
SiamMask \cite{wang_fast_2018}  & SM      & CVPR19                     & S            & N             \\
DiMP \cite{bhat_learning_2019}     & DP      & ICCV19                     & OD           & N             \\
PrDiMP \cite{danelljan_probabilistic_2020}    & PR      & CVPR20                     & OD           & N             \\
SuperDiMP & SD      & 2020                       & OD           & N             \\
KeepTrack \cite{mayer_learning_2021} & KT      & ICCV21                     & OD           & N             \\
KeepTrackFast \cite{mayer_learning_2021} & KF      & ICCV21                     & OD           & N             \\
TransT \cite{chen_transformer_2021}   & TT      & \multicolumn{1}{l}{CVPR21} & S            & Y             \\
TrSiam \cite{wang_transformer_2021}   & TS      & \multicolumn{1}{l}{CVPR21} & S            & Y             \\
TrDiMP \cite{wang_transformer_2021}   & TD      & \multicolumn{1}{l}{CVPR21} & OD           & Y             \\ \bottomrule
\end{tabular}%
}
\caption{Overview of selected trackers, the abbreviations we use to denote them, their source year, general architecture (Siamese-based vs. Online Discriminator based), and an inclusion of a transformer network for additional attention processing. Note that SuperDiMP is a combination of PrDiMP and DiMP provided by Danelljan et al. \cite{danelljan_probabilistic_2020} without further publication, and KeepTrackFast uses slight changes in parameters, but is the same architecture as KeepTrack, and is provided in \cite{mayer_learning_2021}.}
\label{tab:tracker-list}
\end{table}

\subsection{Tracker selection}
Here we discuss which trackers we selected for comparison. We focus on semi-supervised single-object trackers because we intend to use these on vision-based AUVs for tracking of individual marine organisms where large training datasets do not exist. In real-time, semi-supervised object tracking, given a video or live-stream of images, an initial target is determined from a user-specified bounding box in the initial frame. Only based on this information, the tracker must predict the target bounding box on all subsequent frames, in real-time, without further user input.

While semi-supervised trackers have a long history with many types of architectures, performance in all recent benchmarks such as LaSOT, GOT10K, UAV20L, TrackingNet, and more, are all dominated by those based on deep-learned neural networks. We thus focus on those trackers. We selected 13 recent trackers who self-report highest scores on previous benchmarks, discussed in \Cref{sec:related_work}, with varying underlying architectures. As an additional baseline, we include older top-performing trackers as reported by the official LaSOT benchmark. We further select trackers that are capable of running in real-time (above 10fps) on currently available hardware and are used specifically for single-object tracking tasks. As described in \Cref{sec:related_work}, we also focus on two primary architectures, Siamese-based and online discriminative-based trackers. For evaluations we have thus selected: SiamRPN++ \cite{li_siamrpn_2019}, DaSiamRPN \cite{zhu_distractor-aware_2018}, SiamMask \cite{wang_fast_2018}, ECO \cite{danelljan_eco_2017}, ATOM \cite{danelljan_atom_2019}, DiMP \cite{bhat_learning_2019}, PrDiMP \cite{danelljan_probabilistic_2020}, SuperDiMP, which is a combination of DiMP and the regressor from PrDiMP, KeepTrack and KeepTrackFast \cite{mayer_learning_2021}, TransT \cite{chen_transformer_2021}, TrSiam \cite{wang_transformer_2021}, and TrDiMP \cite{wang_transformer_2021}. For abbreviations and properties we consider refer to \Cref{tab:tracker-list}. As is common practice, we run each algorithm as provided \cite{fan_lasot_2020}, because the following reasons: they may require different training strategies, they are sensitive to training settings, and as-is trackers are typically already optimized. There are many subtle differences between each tracker that are difficult to list and enumerate, but broadly we can characterize them as Siamese-based: SiamRPN++, DaSiamRPN, SiamMask, ECO, TransT, and TrSiam, or online-discriminator based: DiMP, PrDiMP, SuperDiMP, KeepTrack, and TrDiMP. All these algorithms utilize pre-trained (using generic datasets) deep convolutional neural networks to perform feature extraction. In most cases, we use the ResNet-50 \cite{he_deep_2016} feature extractor when available, though older networks such as ECO relies on VGG-M \cite{chatfield_return_2014}, and ATOM relies on ResNet-18 \cite{he_deep_2016}. 

In 2017, Vaswani et al. \cite{vaswani_attention_2017} introduced Transformers as a new type of network architecture that produced state-of-the-art results on a variety of neural network related classification and decision making tasks. Only recently however have these innovations been applied into the tracking domain, and are presented in TransT, TrDiMP, and TrSiam. Their performance has not been characterized by a standardized dataset yet, and so we include them here as well. For further details on all trackers, please refer to each respective publication.

\subsection{Results of evaluation and metrics}
\begin{figure}[ht]
    \centering
    \includegraphics[width=0.49\textwidth]{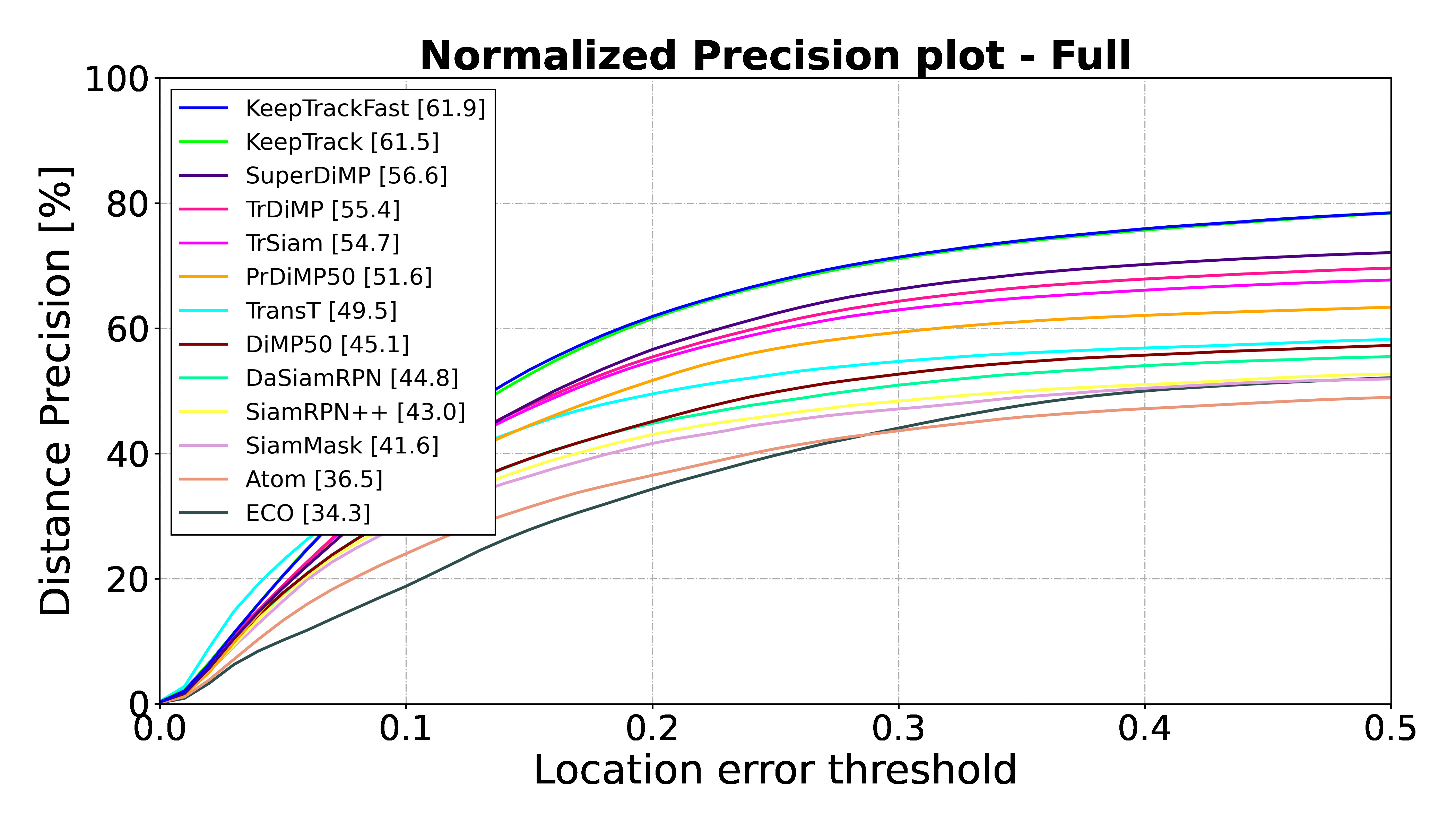}
    \includegraphics[width=0.49\textwidth]{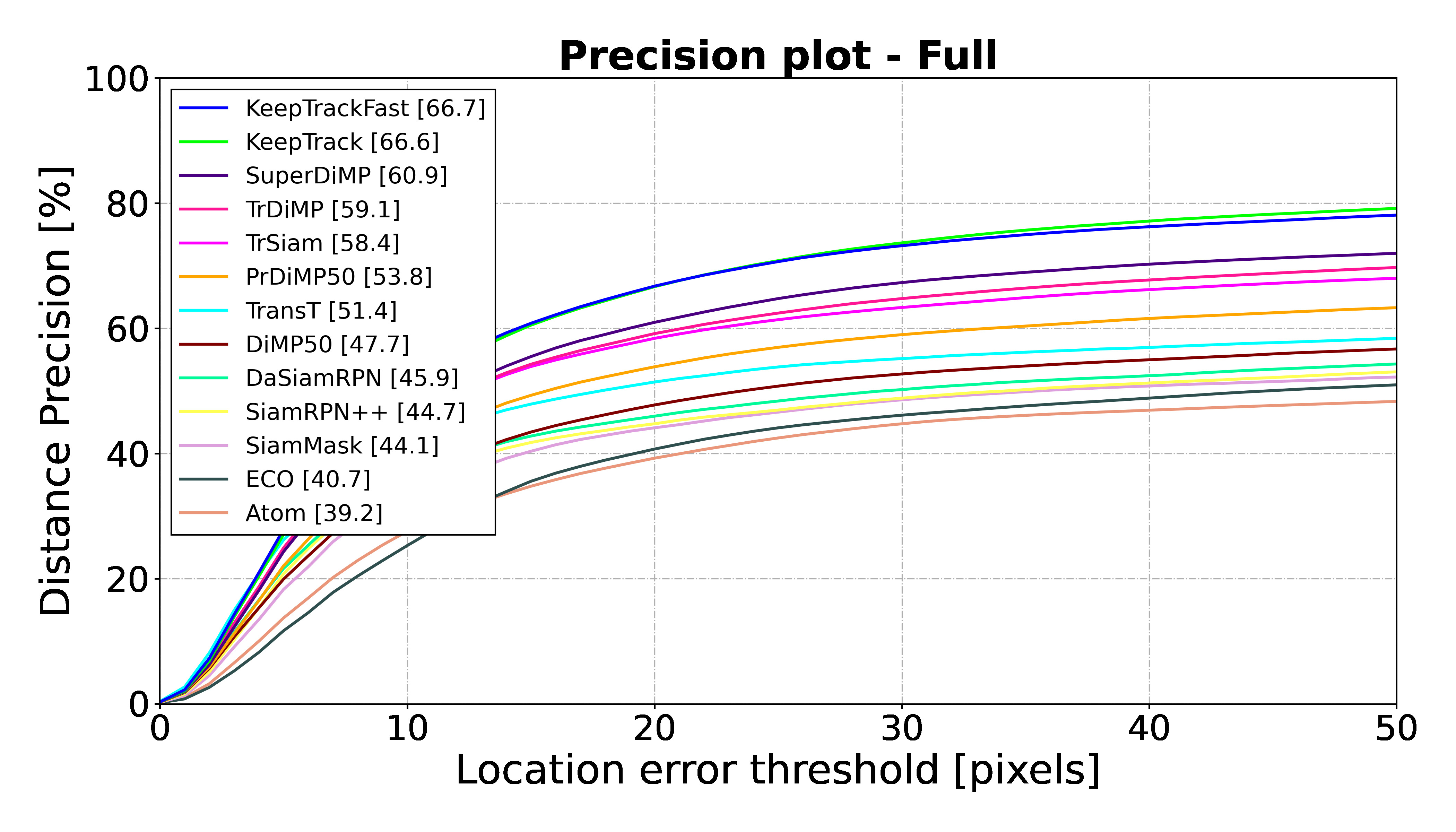}
    \includegraphics[width=0.49\textwidth]{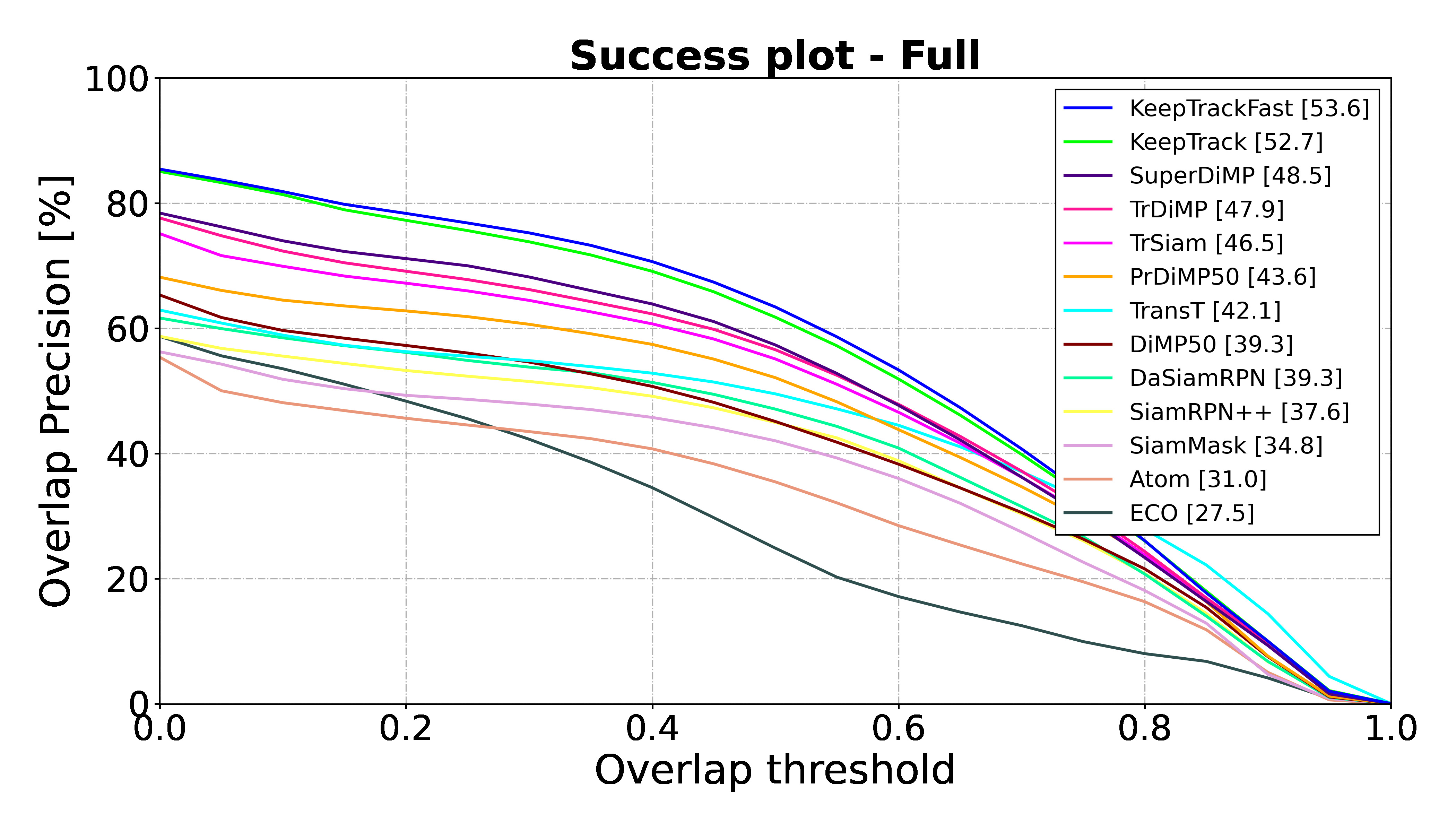}
    \caption{Results of all trackers on the full dataset using success AUC, precision, and normalized precision metrics.}
    \label{fig:full_results}
\end{figure}

We adopt standardized metrics for evaluating state-of-the-art trackers on our benchmark dataset. Namely we consider the (1) \textit{success rate}, (2) \textit{precision}, and \textit{normalized precision} metrics accumulated across all frames in each sequence, and averaged across the entire dataset and over each attribute subset. Many of the trackers are stochastic, and so we average results over 5 runs, as in many other benchmarks \cite{fan_lasot_2020,ferrari_trackingnet_2018-1}. These metrics are well-established and described in detail in 
\cite{ferrari_trackingnet_2018-1}. For completeness, we give brief descriptions of each here. The success rate, or overlap, for each frame and tracker is computed by taking the intersection-over-union (IoU), which is a value between 0 and 1, of the groundtruth bounding box and the predicted bounding box of the tracker. To generate reasonable visualizations and rankings, the IoU is subject to thresholds spanning 0 to 1, a frame is considered a "success" if the IoU exceeds the threshold or not. The percentage of "successful" frames can then be computed for each threshold as shown in \Cref{fig:full_results}. Rankings are then taken by estimating the area under the success curve (AUC). Because some trackers may have reasonably predicted overlapping bounding boxes but focus on the incorrect region of an object, it is standard practice to also consider the \textit{precision} metric as well. Precision is computed by taking the distance of the center of the groundtruth bounding box to the center of the predicted bounding box, this is measured in pixels. For precision, a different threshold, measured using number of pixels distance, is applied to generate meaningful visualizations and rankings. For overall precision rankings in \Cref{fig:full_results}, rather than computing an AUC-like metric for precision, it is standard practice to report the precision for a threshold of 20px, which we do here as well. Since distances are measured in pixels, precision is subject to resolution of the target, and so normalized precision accounts for image resolution. More details about these metrics can be found in \cite{ferrari_trackingnet_2018-1}.

\begin{figure}
    \centering
    \includegraphics[width=0.75\textwidth]{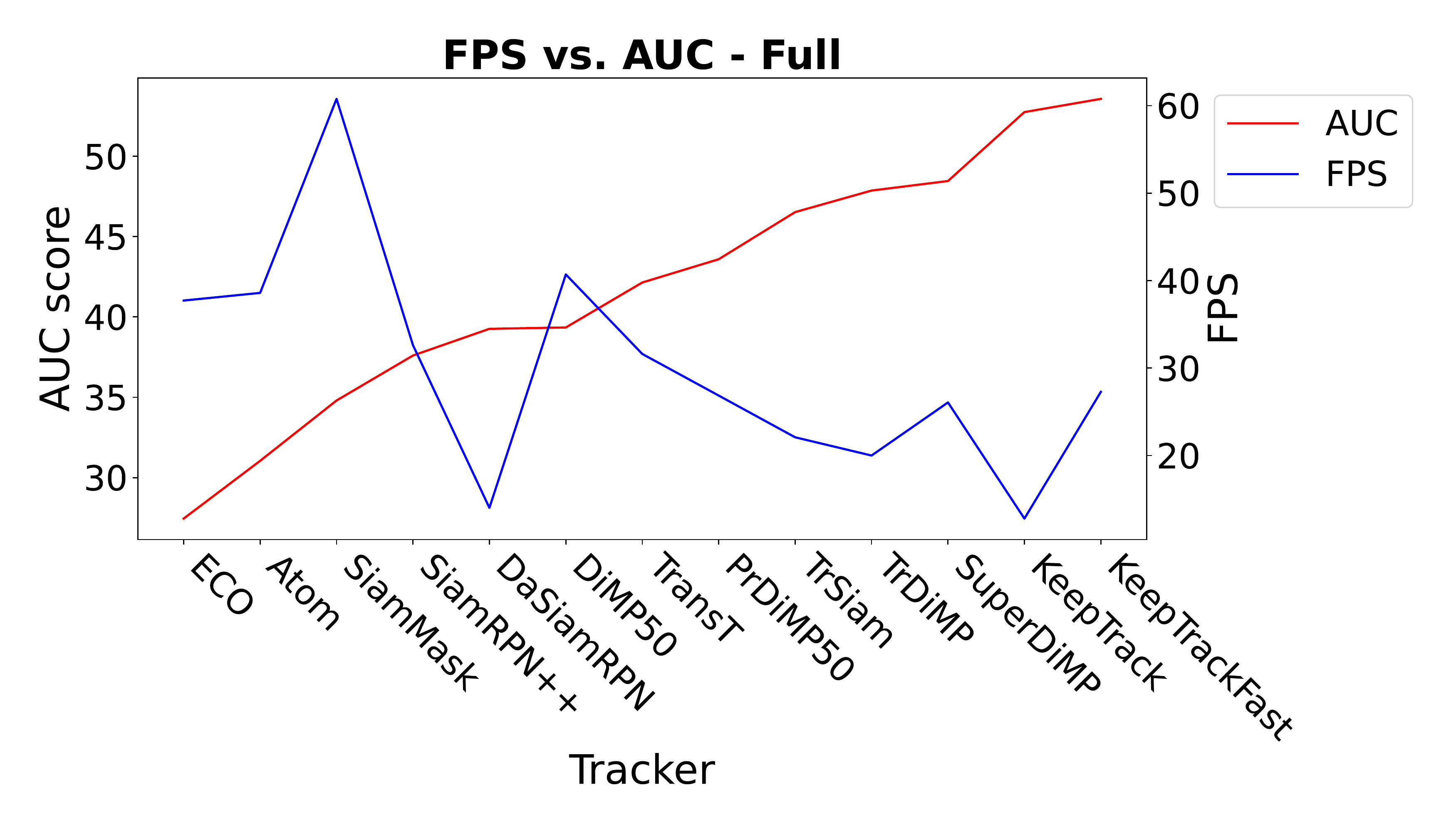}
    \caption{Frame rate vs. AUC over the full VMAT dataset for each tracker. With the exception of KeepTrackFast, SiamMask, and DiMP, most trackers appear to tradeoff speed for accuracy. Results run using a desktop with Nvidia GeForce 1080 GPU, Intel Core i7-6900K CPU, and 64GB RAM.}
    \label{fig:fps}
\end{figure}

Some benchmarks, such as \cite{fan_lasot_2020}, do not report speed, which is a critical evaluation metric if these algorithms are to be used in real-world active tracking systems. We thus ensure to provide a baseline measurement of speed in \Cref{fig:fps}.

\begin{figure}[ht]
    \centering
    \includegraphics[width=0.49\textwidth]{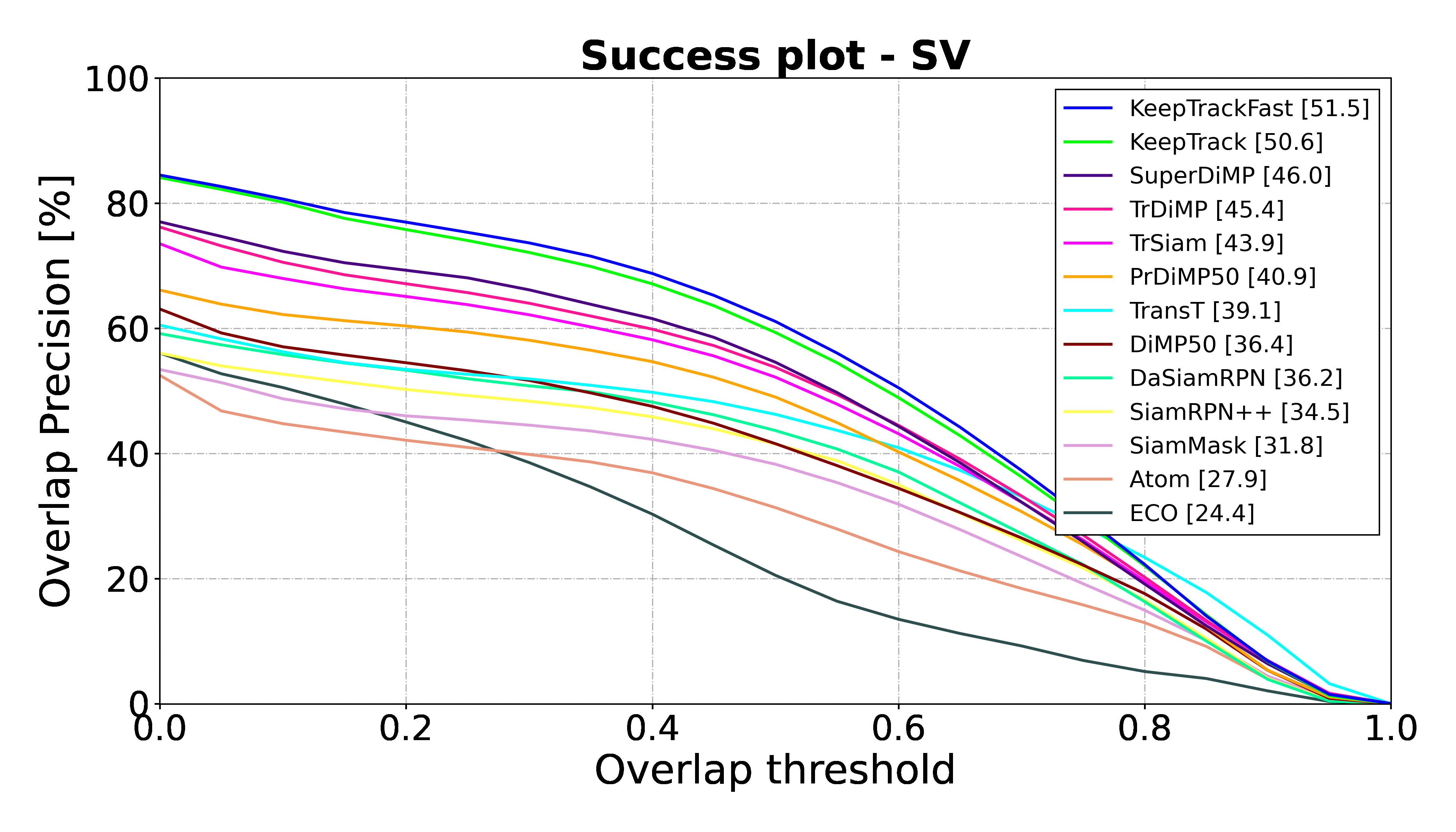}
    \includegraphics[width=0.49\textwidth]{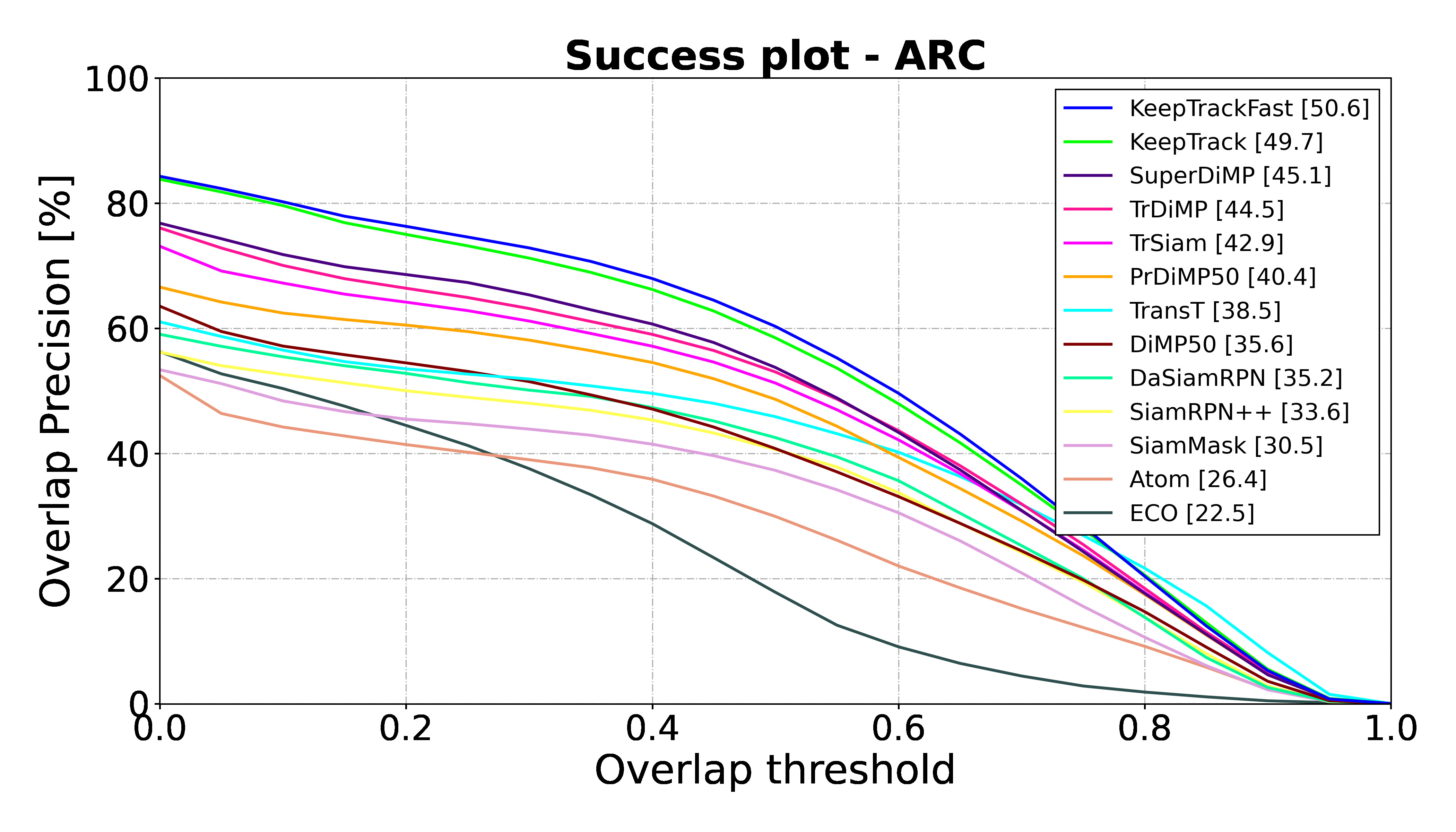}
    \includegraphics[width=0.49\textwidth]{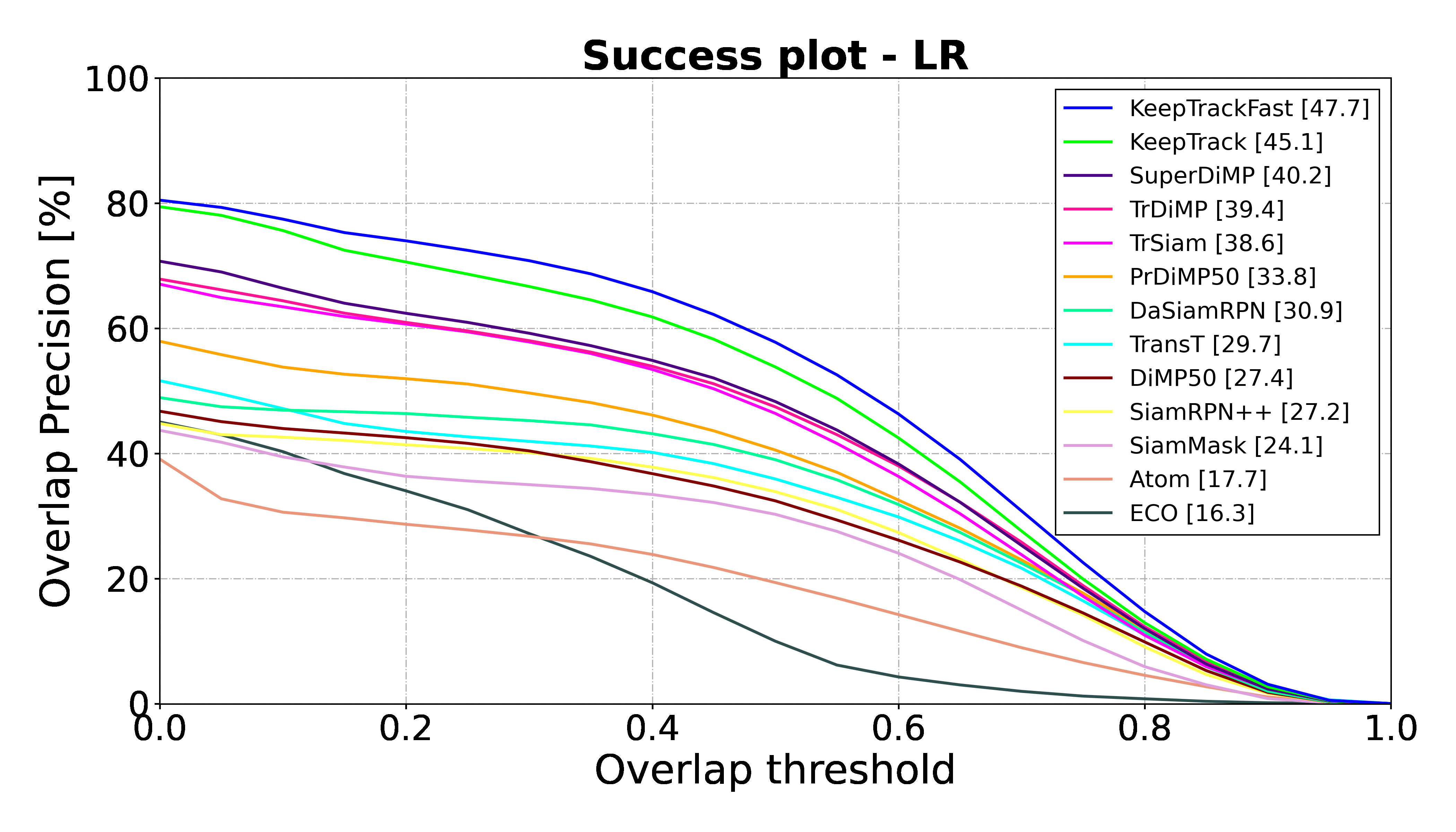}
    \includegraphics[width=0.49\textwidth]{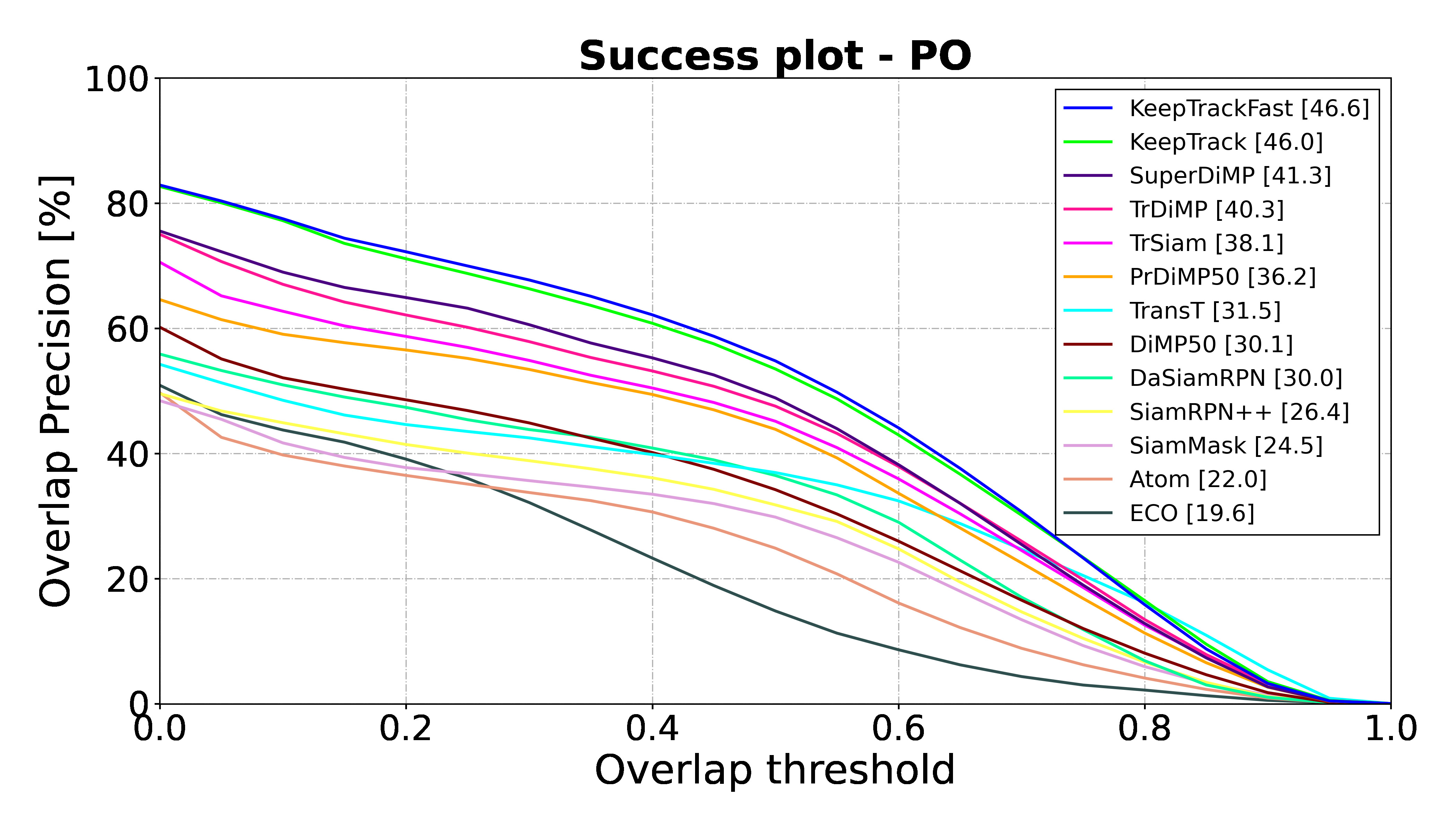}
    \includegraphics[width=0.49\textwidth]{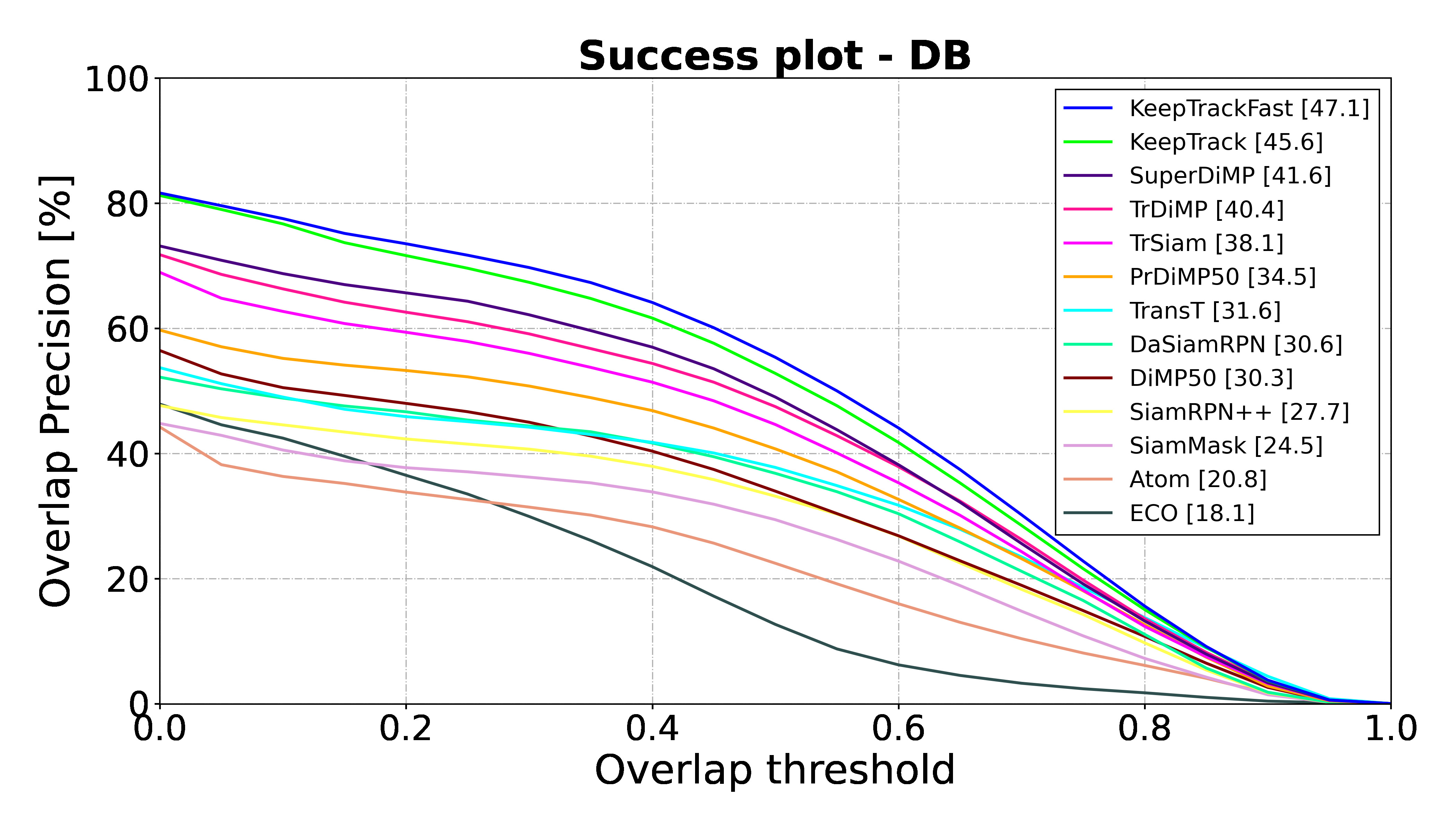}
    \includegraphics[width=0.49\textwidth]{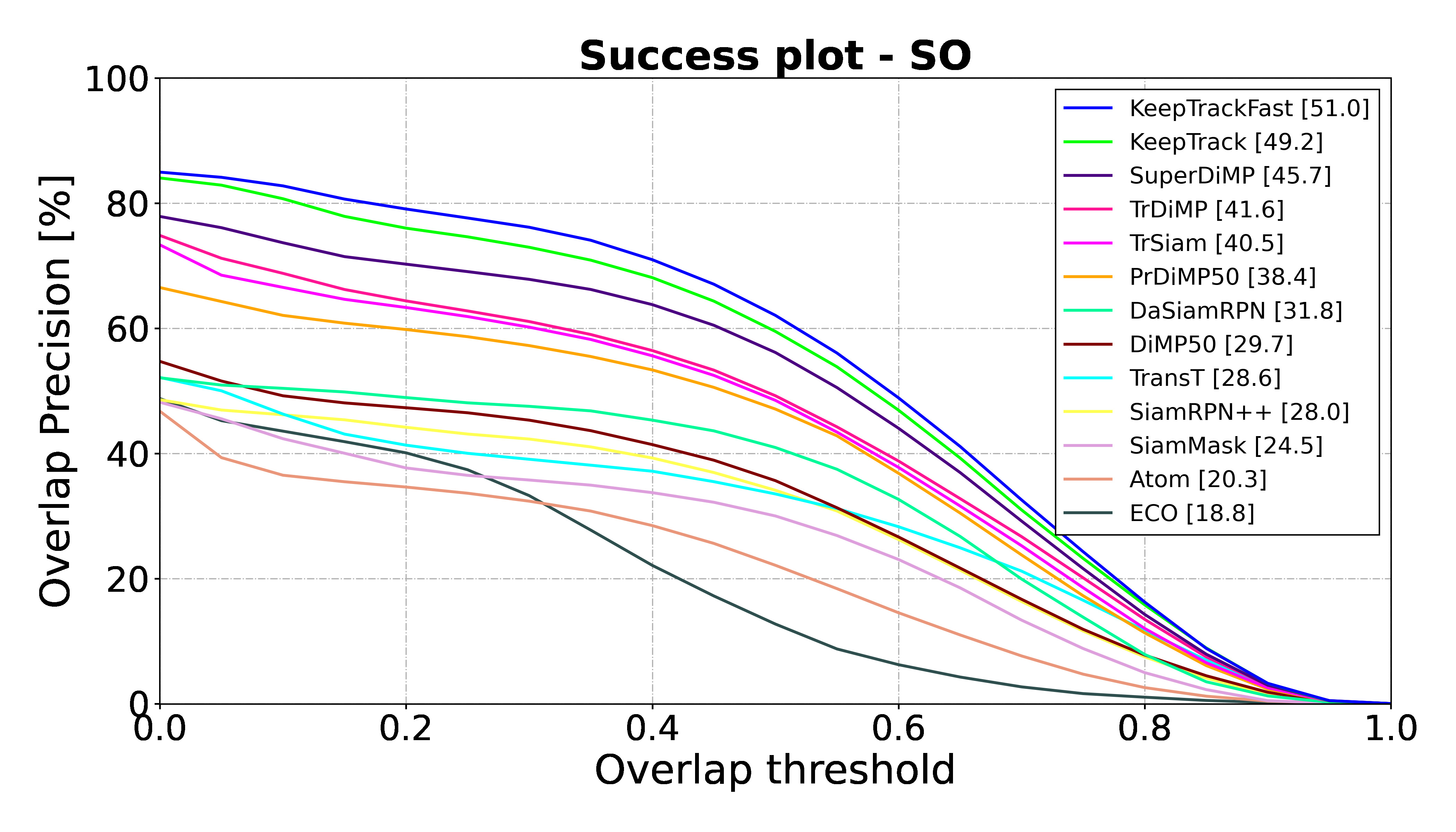}
    \caption{Success results of all trackers on sequences containing the specified generic attributes.}
    \label{fig:attr_results}
\end{figure}

\begin{figure}[ht]
    \centering
    \includegraphics[width=0.49\textwidth]{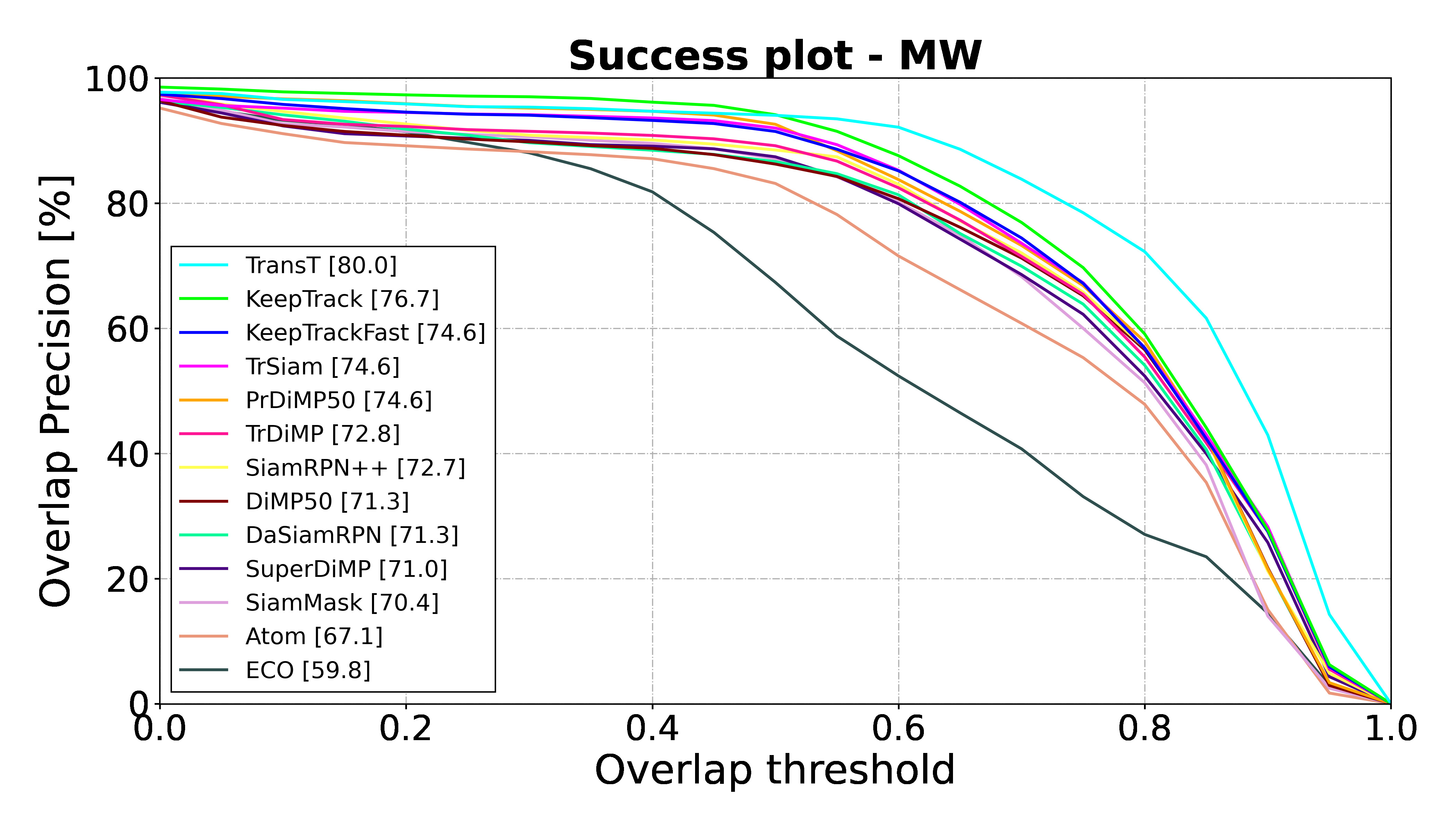}
    \includegraphics[width=0.49\textwidth]{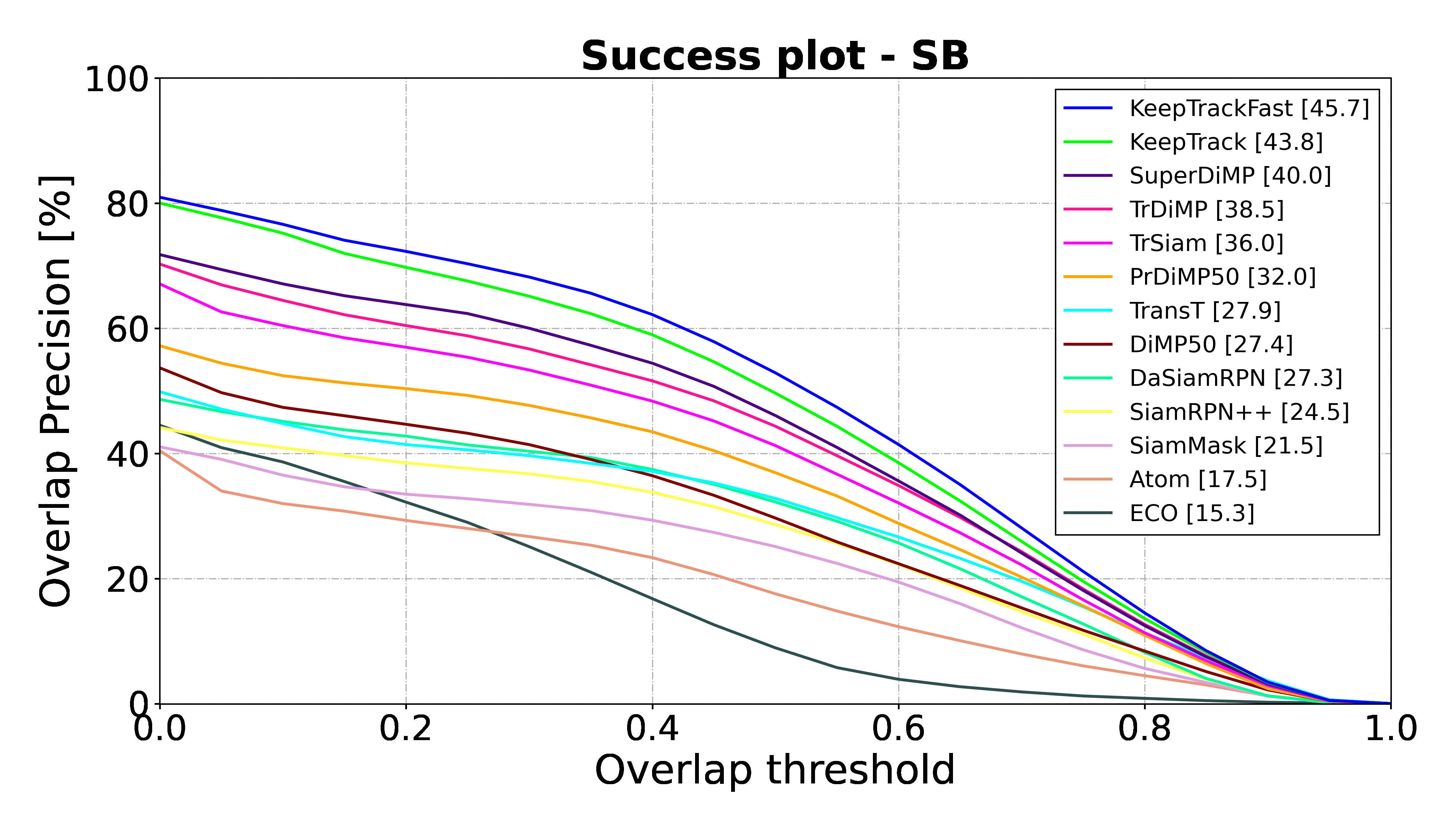}
    \includegraphics[width=0.49\textwidth]{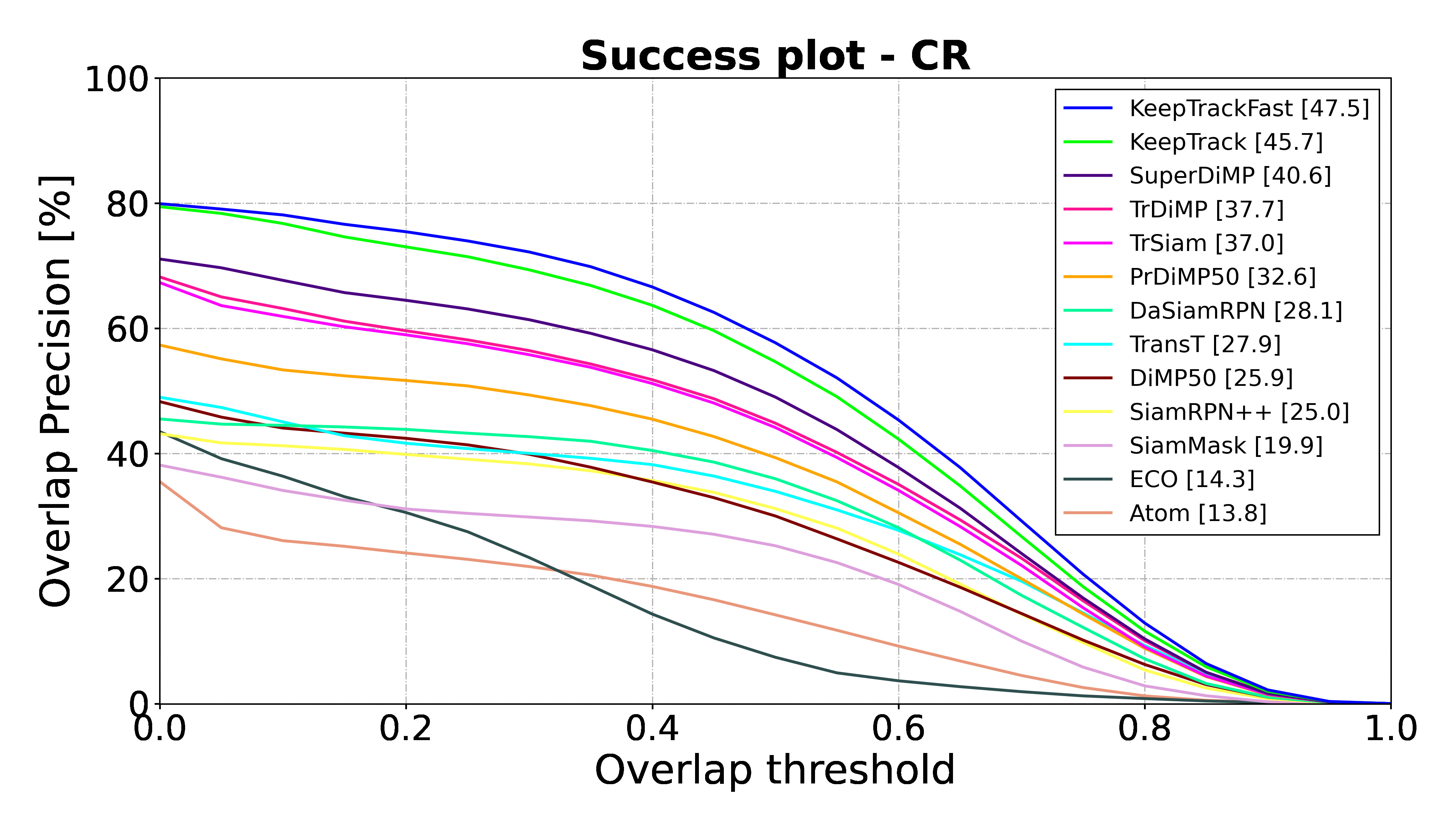}
    \includegraphics[width=0.49\textwidth]{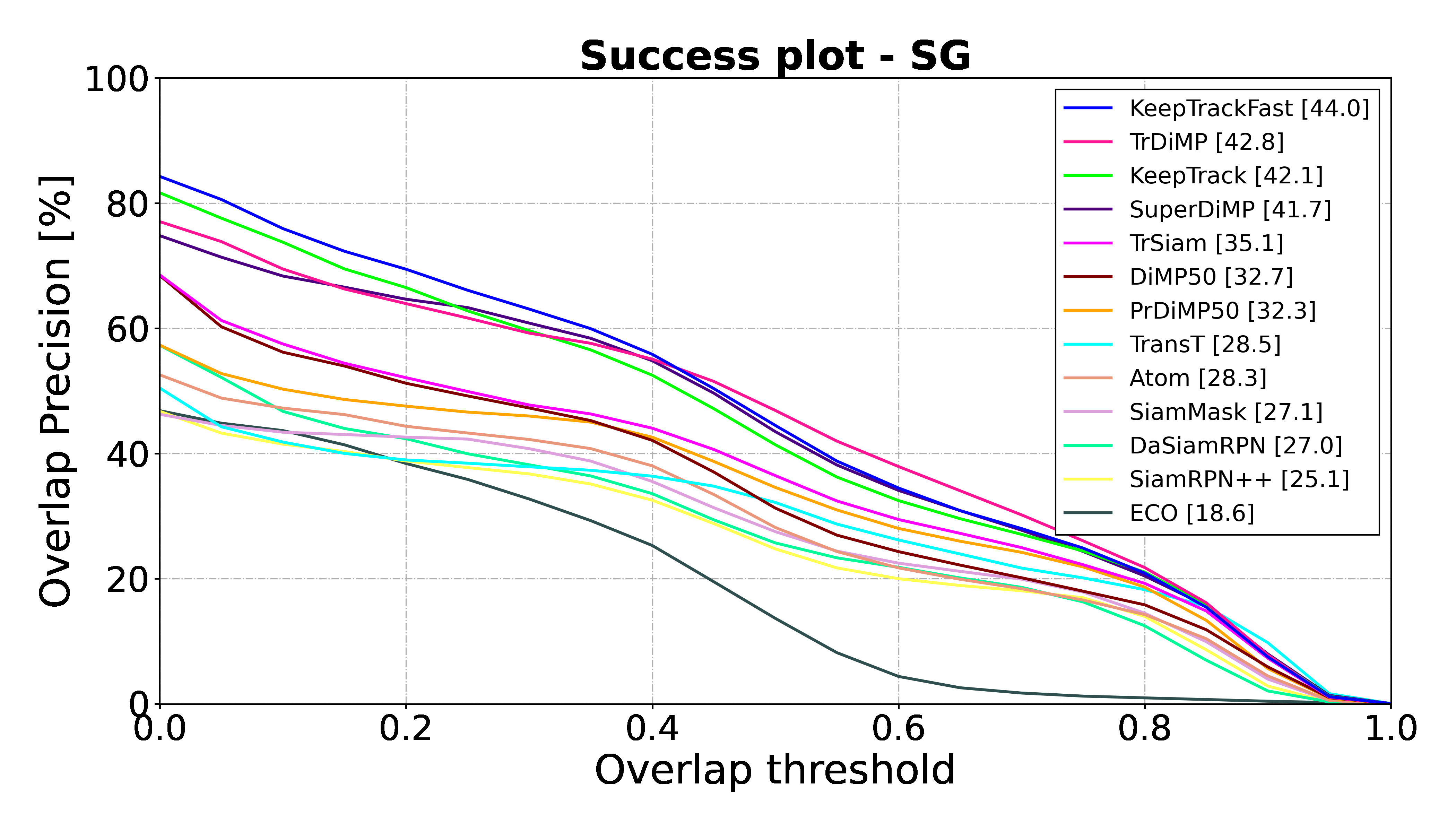}
    \includegraphics[width=0.49\textwidth]{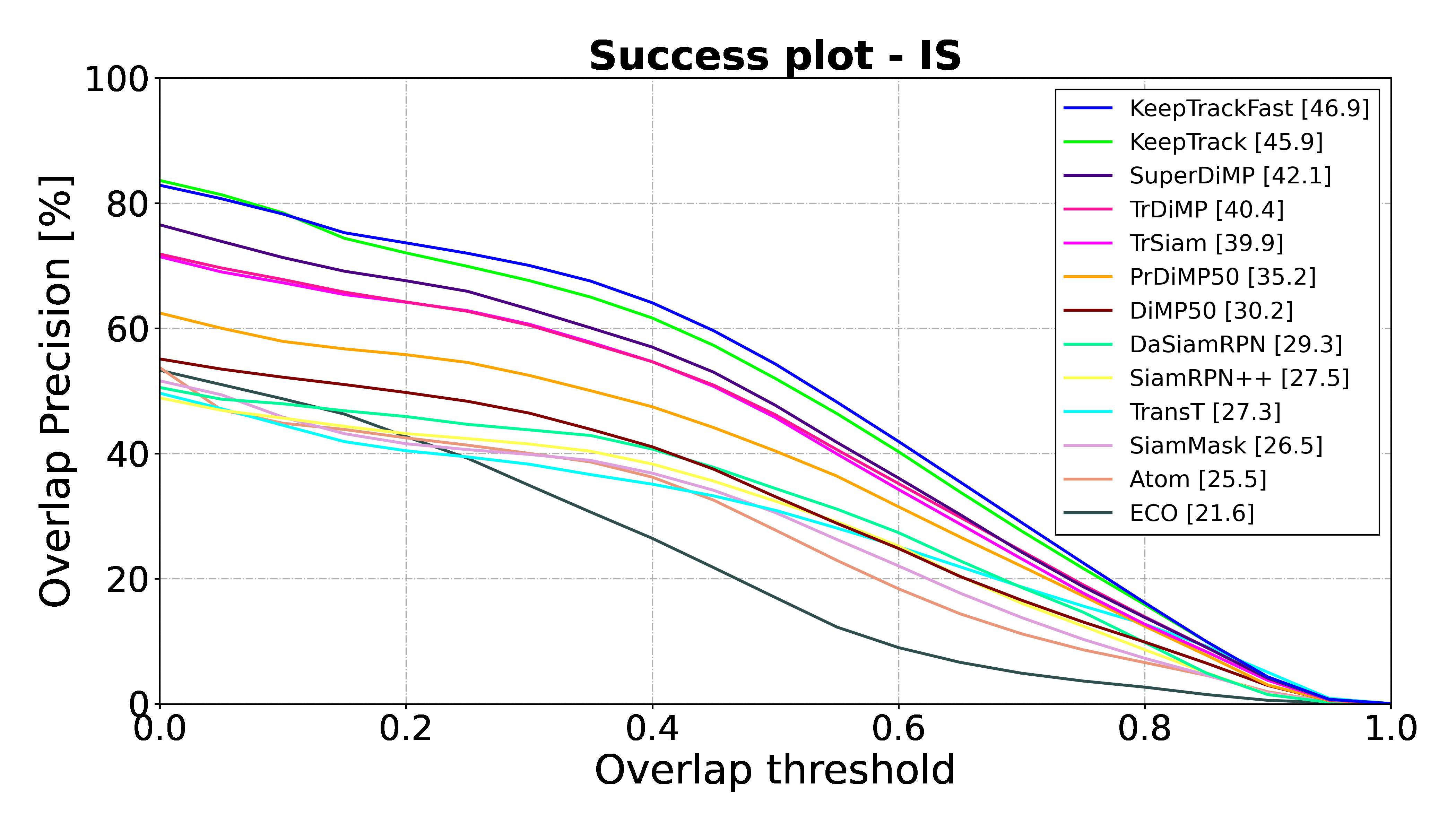}
    \includegraphics[width=0.49\textwidth]{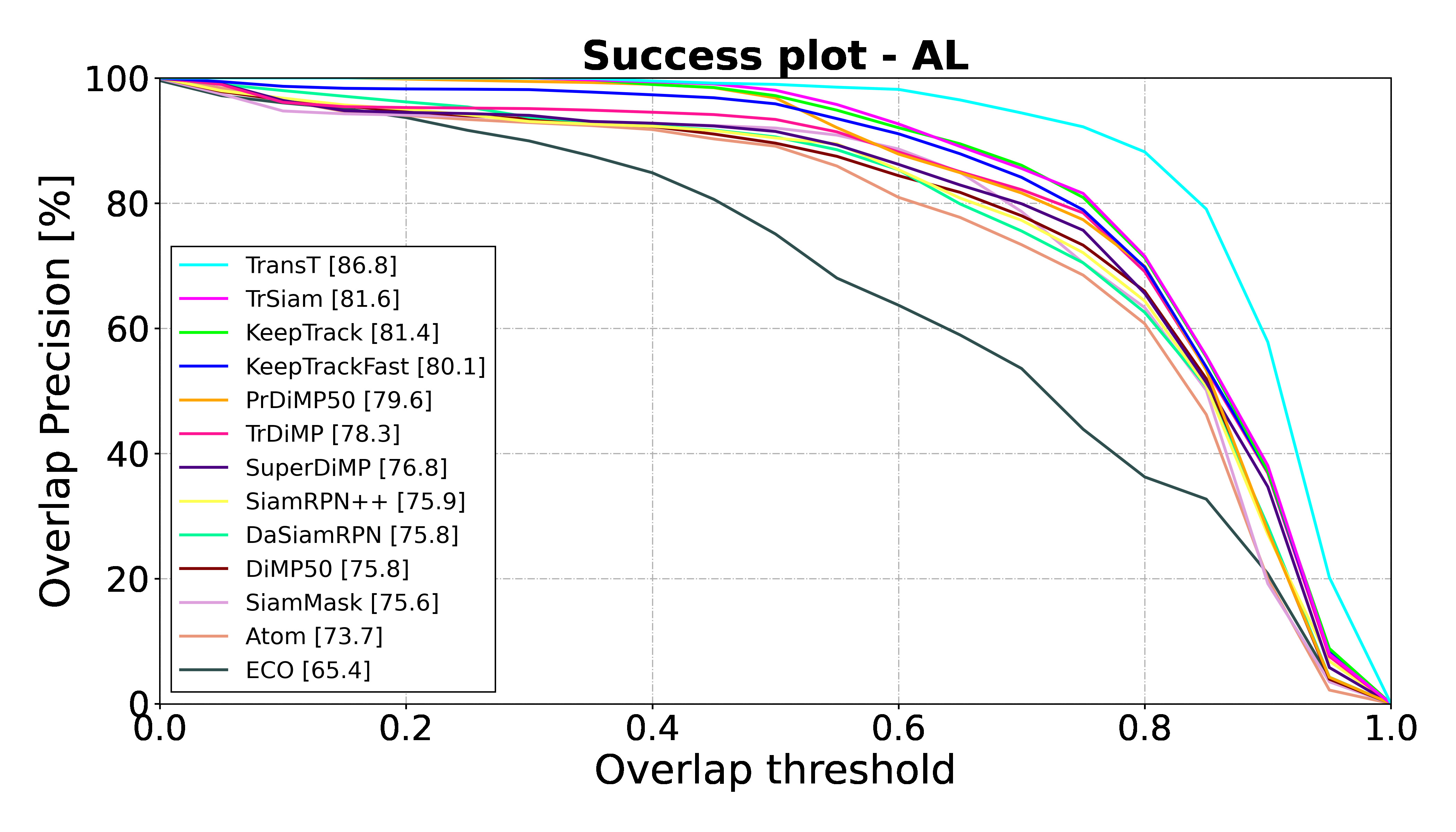}
    \includegraphics[width=0.49\textwidth]{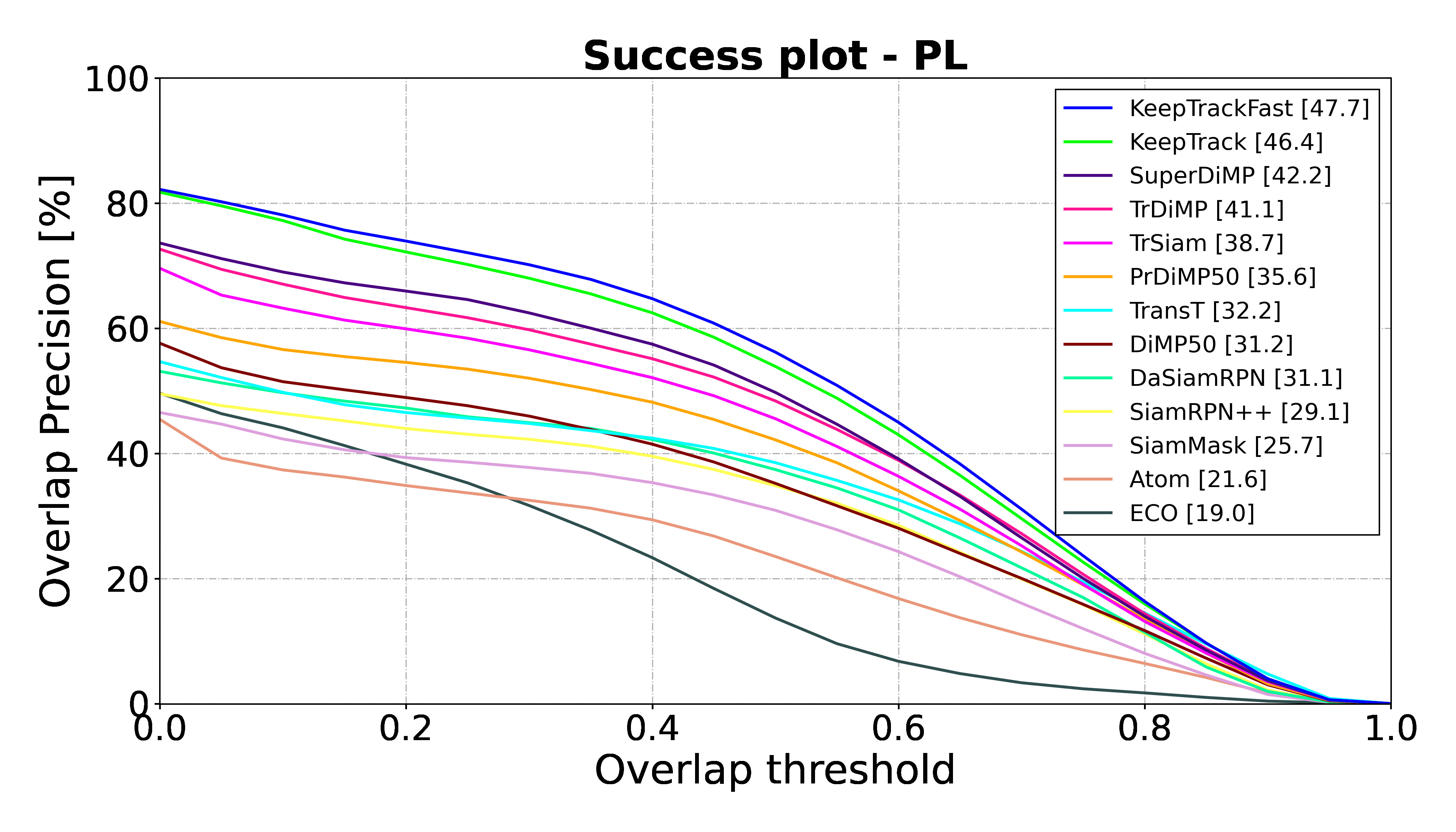}
    \caption{Success results of all trackers on sequences containing the specified environmental attributes.}
    \label{fig:env_attr_results}
\end{figure}

The results for the full dataset can be found in \Cref{fig:full_results}, results along standard attributes are in \Cref{fig:attr_results}, and results along the unique underwater attributes are in \Cref{fig:env_attr_results}. To enable consistent comparisons, especially related to speed, all results are run on the same desktop with a Nvidia GeForce 1080 GPU, Intel Core i7-6900K CPU, and 64GB RAM.

\begin{figure}[ht!]
    \centering
    \includegraphics[width=\textwidth]{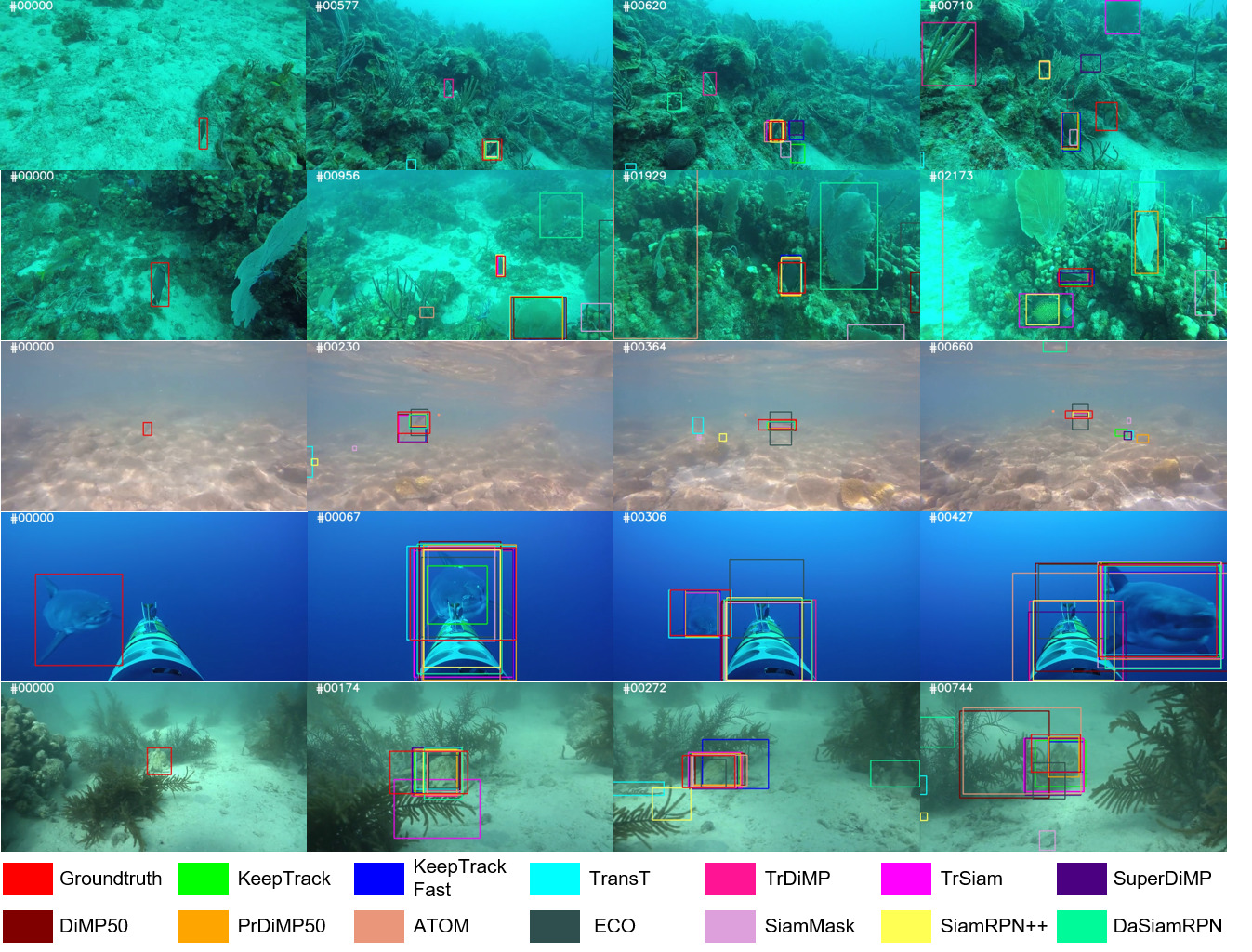}
    \caption{Sample representative results on difficult sequences of a bluetang, angelfish, squid, shark, and octopus (from top to bottom).}
    \label{fig:qual_results}
\end{figure}

\section{Real-world experiments: tracking in the wild using an AUV}
\label{sec:stingray}

\begin{figure}[ht!]
    \centering
    \includegraphics[width=\textwidth]{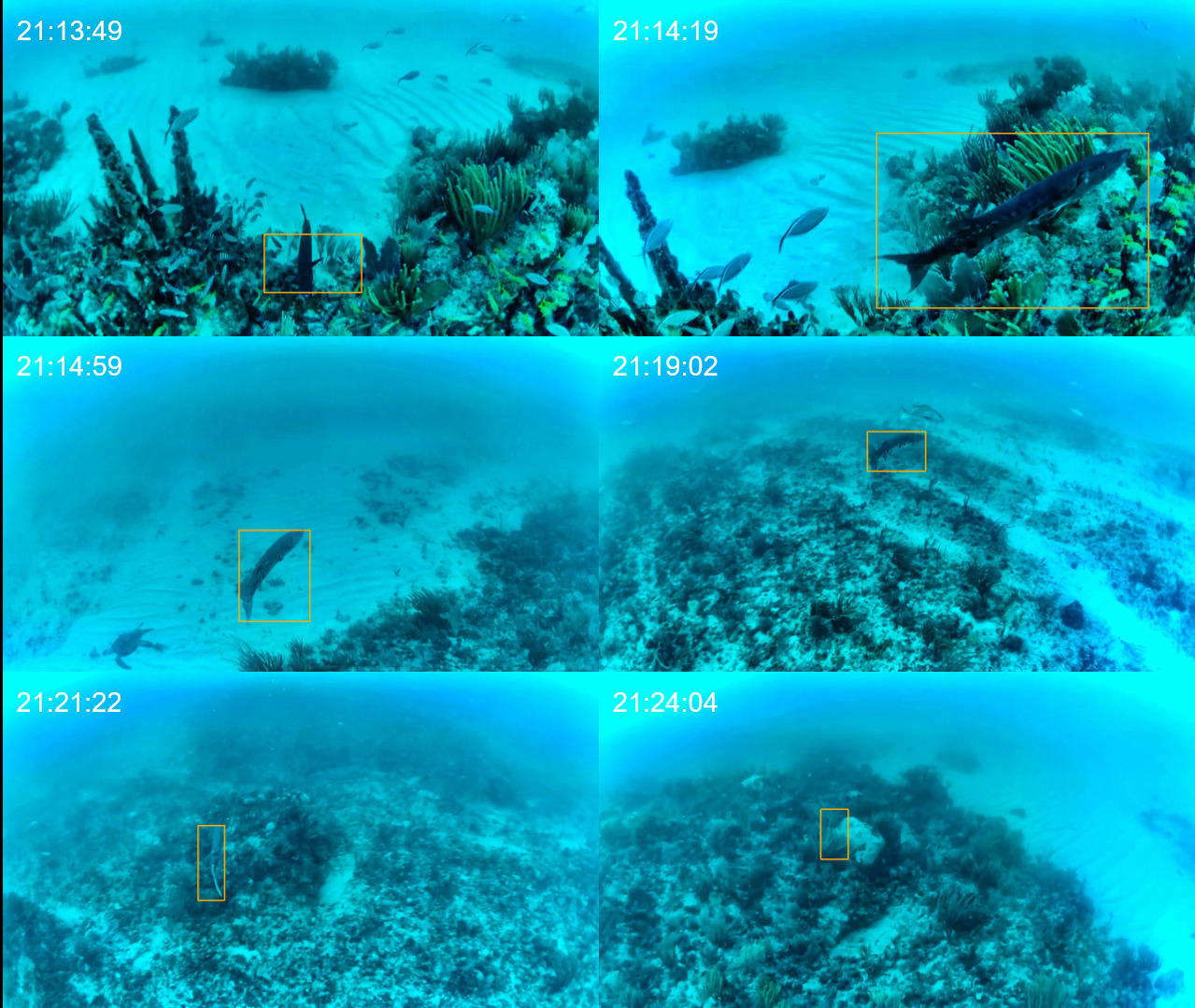}
    \caption{We spontaneously and mostly autonomously tracked a barracuda at Joel's Shoal, St. John, U.S. Virgin Islands, USA on Nov. 3rd, 2022, for roughly 10 minutes using a semi-supervised visual tracker on an AUV. Tracking was performed at roughly 7-9fps at 720p resolution on-board a Nvidia Jetson Xavier using KeepTrackFast \cite{mayer_learning_2021}. Images are taken from the AUV perspective. The UTC time is listed in the upper left of each image. The provided bounding box is shown in the 1st frame, it is not well-centered because the operator had to click-and-drag the box while on-board an oscillating boat, and the rest are automatically generated during tracking and used to visual servo the AUV. In the last frame, the barracuda is actually a few pixels to the upper right of the predicted bounding box, which highlights in extraordinary challenges of this environment, as even a human cannot separate the barracuda from the background, even though it is minimally occluded. The AUV loses the barracuda here and we ended the track. For the full track video, please refer to supplementary video S1.}
    \label{fig:barry}
\end{figure}

\begin{figure}
    \centering
    \includegraphics[width=\textwidth]{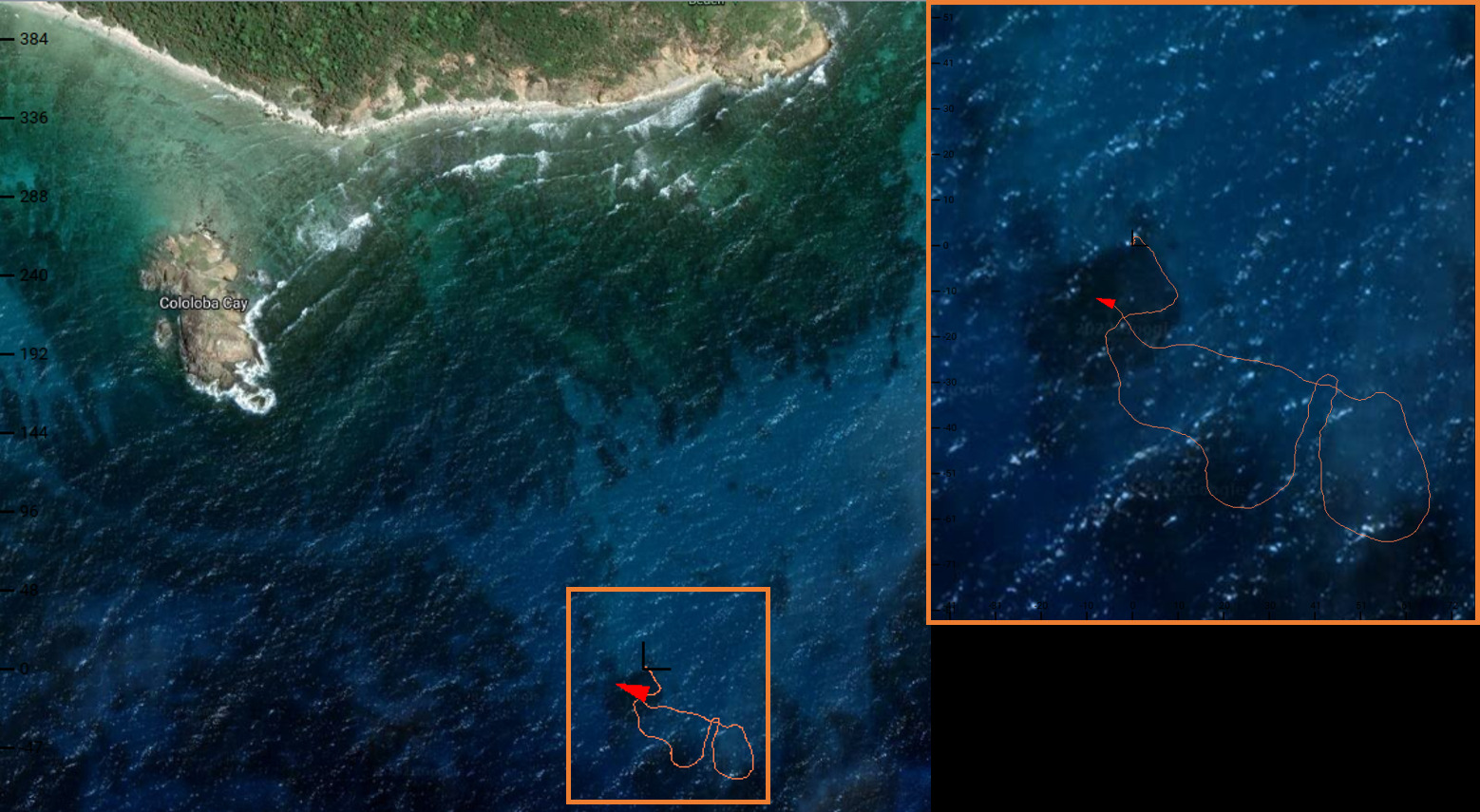}
    \includegraphics[width=\textwidth]{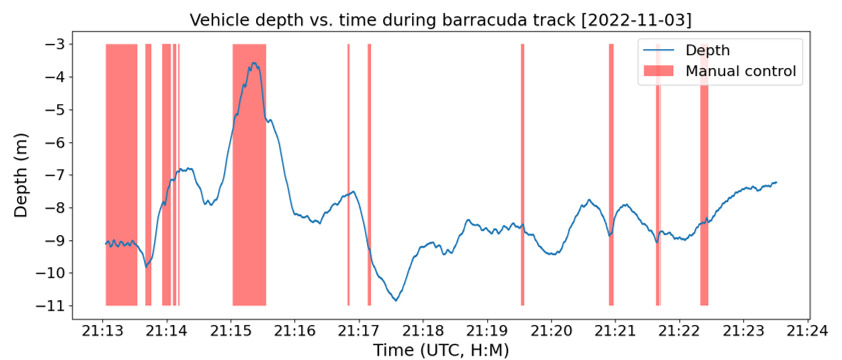}
    \caption{Here we show the full 3D trajectory estimates of the barracuda track from \Cref{fig:barry}, which swam roughly 100 meters out and back. On top is the AUV real-time estimate of its trajectory, achieved via dead-reckoning by integrating DVL measurements, as it follows the barracuda (which required manual post-processing to align it with the GPS map). Finally, on the bottom, we show the vehicle depths during tracking, along with red highlights indicating when the KeepTrackFast algorithm was unable to stay focussed on the correct target and the operator needed to briefly, manually control the vehicle until KeepTrackFast was able to re-acquire the target. The beginning section required manual control due to the initial training time of KeepTrackFast, and intermittent areas when the barracuda resembled the corals underneath, similar to the last frame in \Cref{fig:barry}. Note that these position estimates can serve as proxies for animal positions and velocities. Using stereo, we can generate true animal position and velocity estimates, but is out of the scope of this paper.}
    \label{fig:barry_trajectory}
\end{figure}

\begin{figure}
    \centering
    \includegraphics[width=\textwidth]{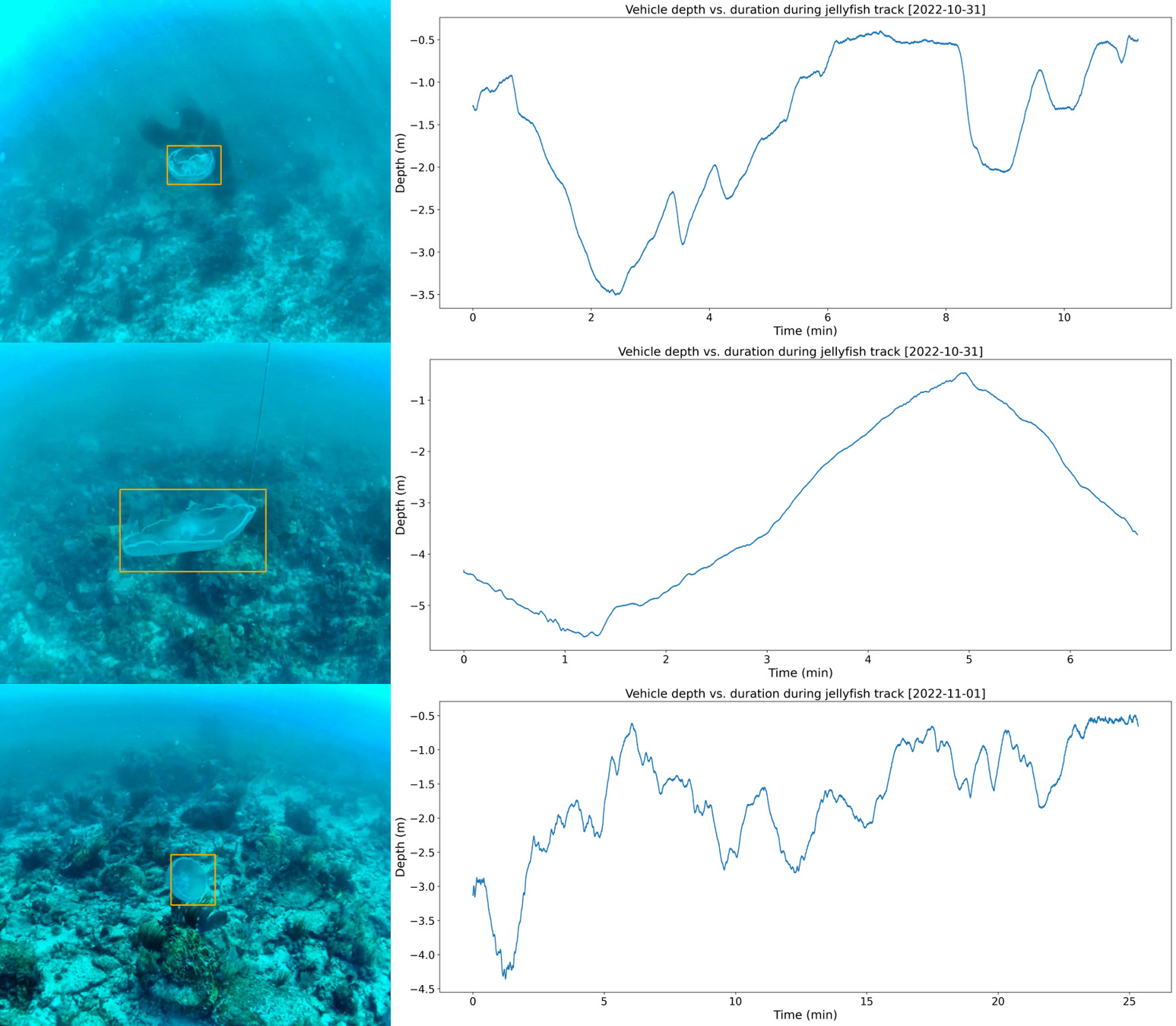}
    \caption{We tracked 3 separate jellyfish at 2 different reefs in St. John, USVI, USA. From top to bottom they were at Tektite, Booby Rock, and Booby Rock for roughly 12, 7, and 25 minutes respectively. Once the initial bounding box was specified, the remainder of the tracks were completely autonomous. The tracks ended because the vehicle was tethered and the jellyfish drifted beyond the tether range. The vehicle depth profiles are shown on the right, which can approximate the depth of the jellyfish, and their initial bounding boxes and images are on the left, which have been cropped for space. Videos of the jellyfish tracks are available in the supplementary materials S2-S4, respectively.}
    \label{fig:jelly}
\end{figure}

\begin{figure}
    \centering
    \includegraphics[width=\textwidth]{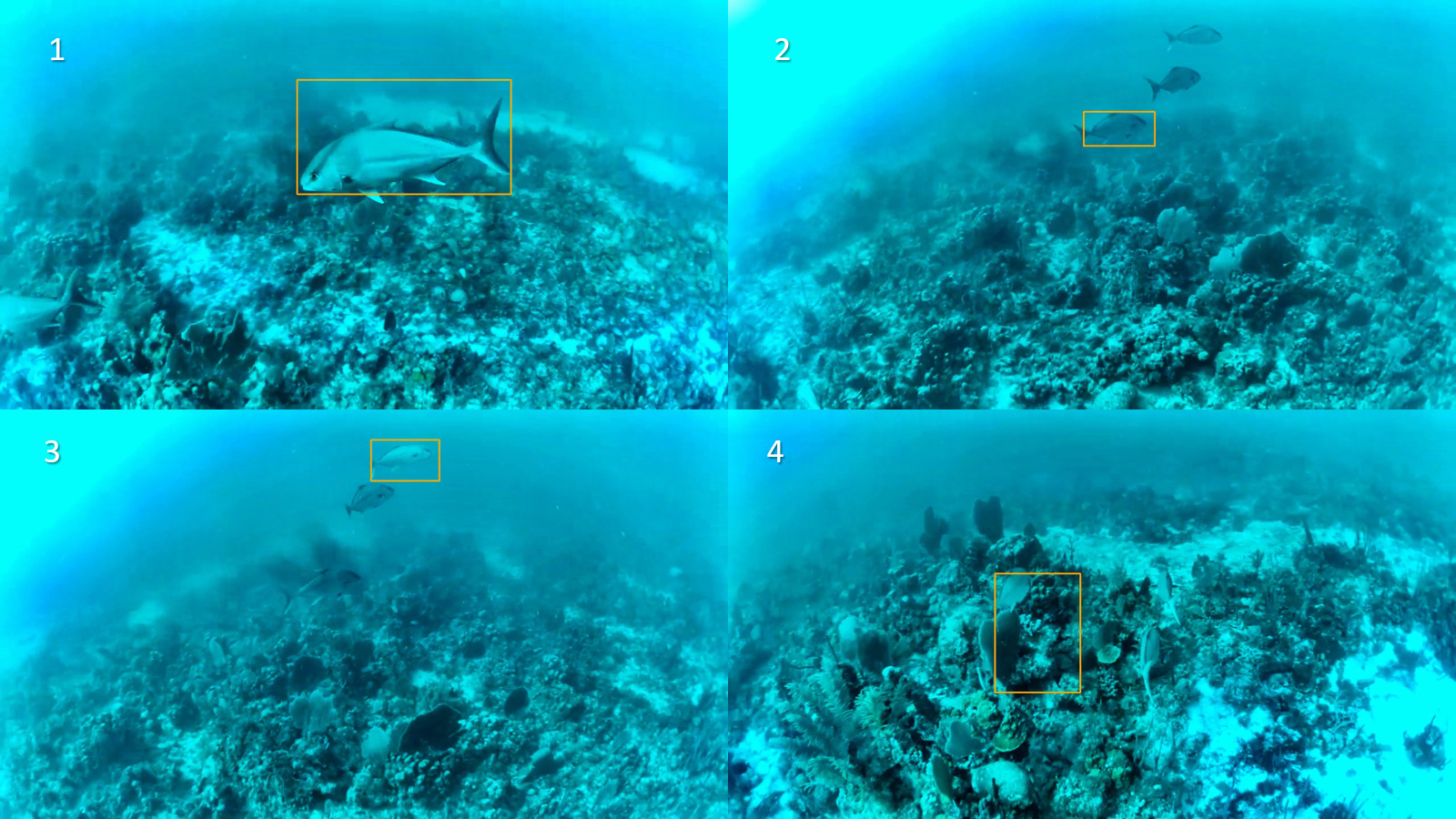}
    \caption{We autonomously track a type of jack that was swimming with two others. KeepTrackFast oscillated between the three, even though it is designed to minimize these confusions. This also caused the vehicle to jerk between all three, and frequently lose track. The issue is especially highlighted when the fish turn to face away from the camera, as in the 4th frame. The thin profile of the fish and homogeneous colors from color absorption makes the fish difficult to distinguish from the background and surrounding corals. These situations trigger KeepTrackFast to oscillate even more between targets and occasionally background regions. Refer to the supplementary material S5 for the jack track video.}
    \label{fig:jack}
\end{figure}

Solely evaluating performance of semi-supervised trackers on pre-recorded videos is insufficient to fully estimate their benefits and issues when operating in real-world conditions. Pre-recorded videos mask issues where real-time decision-making on the AUVs can cause additional problems. For instance, if the tracker focuses on the wrong object even momentarily, it could cause the entire platform to further point away from the target. This creates a positive feedback loop that likely results in catastrophic failure of the tracker. In pre-recorded videos however, trackers can often re-identify targets because it is likely they will re-center in the video eventually given the nature of the recording. Thus, in order to better understand in-situ performance, we deployed a semi-supervised visual tracker to actively control a real AUV and track a marine animal in the wild. We describe our system implementation and experiment details below and provide some qualitative evaluation of the field trials afterward.

\subsection{System Overview}

Our goal was to test semi-supervised tracking of a complex organism in-situ and in real-time on an AUV. To do this we used a custom AUV developed at the Woods Hole Oceanographic Institution, known as CUREE \cite{girdhar_icra_2023}, rated to 100m, equipped with a camera, a high-bandwidth connection to an on-board Nvidia Jetson Xavier which had direct control of the thrusters over a ROS interface \cite{noauthor_ros_nodate}. All image processing and control action selection was performed on the AUV itself. The vehicle is equipped with 3 wide-angle cameras looking 30-degrees down from the horizon. The middle camera provides a 720p RGB stream at 15fps, while the two side cameras provide 1080p monochrome streams and can be used for stereo.
 We used KeepTrackFast because it was both the best performing algorithm on VMAT while also able to run at roughly 7-9fps continuously on the edge device Nvidia Jetson Xavier, which is at the bare minimum speed to enable real-time feedback control. Note that this device is significantly less powerful than the system used in \Cref{fig:fps}, but is necessary for use on smaller autonomous vehicles. This on-board computing limitation is also why the resolution and speed of the video processing was low. The camera is mounted at a 30-degree downward angle from the horizon in order to better track animals along the sea-floor, which is also typically a more complex environment. The AUV is also equipped with a WaterLinked DVL A50, or a \textit{Dopper Velocity Log}, which provides capabilities to accurately estimate 3D vehicle velocities and altitudes, which can in turn be integrated to estimate positions via \textit{dead-reckoning}. Finally, the AUV has 6-thrusters capable of full 6-DOF control.

\subsection{Real-world deployment details}

We deployed our system in parallel to our dataset collection efforts in the U.S. Virgin Islands in October 2022. A human operator was stationed on a boat and was tethered to the AUV. The operator had a high-bandwidth tether connection to the AUV in order to perform remote control of the vehicle for the initial target search and specification step of the experiment. This connection also provides the operator with a live stream of the video from the AUV. Once a target is identified by the operator, they draw a bounding box over the target using a GUI developed for this task. After the bounding box is specified, the AUV goes into a fully autonomous tracking mode. While this may be considered an ROV setup, once the initial bounding box has been specified, the tether can be disconnected, and the goal is for these algorithms to be adopted on full AUVs.

In fully autonomous mode, the AUV performs closed-loop visual servoing control using the bounding boxes provided by KeepTrackFast. Because of the monocular setting, and because there is no known information about the target a priori, we cannot compute a target size or distance estimate. Consequently, we command the vehicle to maintain constant width of the bounding box. Qualitatively, we found this metric to be more stable than height or area. The resultant control loop guides the robot closer to the target if the width of the bounding box grows, and vice versa. The AUV also yaws accordingly to center the target in its field of view. If the center is above or below, the vehicle rises or sinks in the water column. All these parameters are controlled via a series of PID loops running at least at 9Hz, which is the speed of KeepTrackFast running on the Jetson Xavier. For additional safety, to prevent damaging coral reefs, using the DVL's measurement of altitude, we disallow the vehicle from moving below 0.5m to 1m from the bottom.

\subsection{Results of demonstrations}

We were able to successfully track several organisms, including a barracuda (\Cref{fig:barry}), multiple jellyfish (\Cref{fig:jelly}), and some smaller fish like jacks (\Cref{fig:jack}). 

The barracuda was tracked for nearly 10 minutes across a trajectory of roughly 100 meters in length out and back. The barracuda swam from an initial point next to a particular species of coral, known as \textit{dendrogyra}, had multiple encounters with other organisms along its track, and eventually returned to the same coral before the AUV lost track. The AUV was able to autonomously follow the barracuda for the majority of the track, however, there were short instances where the human operator needed to manually control the AUV when KeepTrackFast lost the target. Overall, the tracker was able to maintain track of the animal across a wide range of terrains, as it swam over coral, sand, seagrass, and nearby several other fish, turtles, and other organisms. However, at around 10 minutes, it swam near coral that also resembled it, and KeepTrackFast and the human operator as well were unable to see it afterwards. Because of the DVL and depth sensor on-board the AUV, we are able to estimate a full 3D trajectory of the vehicle in real-time, which can serve as a proxy of the trajectory of the barracuda. The AUV also has stereovision, so with further processing, we would be able to recover a more accurate 3D trajectory of the barracuda itself by computing the 3D offset of the barracuda from the vehicle, however, that is out of scope for this paper. The resulting trajectory estimates, and time periods when human intervention was required, are shown in \Cref{fig:barry_trajectory}. We needed to manually align the vehicle compass, due to a compass miscalibration while we were tracking the barracuda, in order to post-process the map overlay in \Cref{fig:barry_trajectory}. 

We further initially tested the capability of these algorithms to replicate results such as \cite{yoerger_hybrid_2021, katija_visual_nodate-1} to track jellyfish in more visually complex scenarios. To do this, we tracked multiple jellyfish, the resulting vehicle depth plots, which again, roughly estimate the depth profiles of the jellyfish, are shown \Cref{fig:jelly}. While we were unable to test in situations with large clusters of jellyfish as in \cite{katija_visual_nodate-1}, we were still able to track jellyfish in more visually complex scenarios for reasonable time lengths. The tracks ended because the vehicle is tethered to the boat of the human operator, which is in turned moored, and once the jellyfish reached beyond the length of the tether it was no longer possible to continue. However, in the case of the jellyfish tracks, once the initial bounding box was specified, the entire track was fully autonomous.

Finally, we attempted tracks of many smaller organisms, of which we show one demonstration of a track of a jack in a small school of three in \Cref{fig:jack}. In this track, three jacks were swimming together and we commanded the vehicle to track one. We were able to track at least one jack for several minutes, though the human operator needed to intervene for a few seconds several times in cases where the tracker or the vehicle was unable to maintain the track because either the tracker got confused and tracked a coral, or the vehicle itself was physically slower.

\section{Discussion}
\label{sec:discussion}

\subsection{Tracker evaluation}
\label{sec:tracker_discussion}
In \Cref{fig:full_results}, we see that the KeepTrack algorithm (KeepTrackFast is the same underlying algorithm, just slightly different parameters to improve speed) the best performing algorithm over the entire dataset and, in \Cref{fig:qual_results} and \Cref{fig:env_attr_results}, along most attributes by a fairly wide margin, having a 53.6 and 52.7 success overall and 66.7 and 66.6 precision overall for KeepTrackFast and KeepTrack, respectively. The next best performing is SuperDiMP, achieving 48.5 and 60.9 AUC and precision scores respectively. This makes sense because KeepTrack is based on SuperDiMP, and is specifically designed to handle distracting objects and backgrounds, which make up over 35\% and 70\% of the dataset. Surprisingly, transformer-based networks (TrDiMP, TrSiam, and TransT), which are mostly based on SuperDiMP, perform marginally worse than SuperDiMP in most cases. The only cases where these seem to exceed performance is in the midwater and active lighting results, where all trackers performed exceptionally well, likely because the backgrounds are much less challenging relative to the targets in our dataset. Also, online discriminative style trackers tend to perform better than Siamese-based ones.

In terms of the standardized attributes shown in \Cref{fig:attr_results}, the rankings are mostly stable, suggesting that some of these attributes are either heavily correlated or that certain innovations seem confusing across multiple of these dimensions, making them still difficult to analyse fully. However, it is clear that low-resolution frames are still a significant concern, with the best performing tracker only achieving 47.7 success rate. These are exacerbated in many marine animals, such as several fish species or in the case of darting octopuses, where their bodies become extremely narrow and difficult to track purely from appearance. We believe these situations can be mitigated with a more probabilistic handling of search area, which none of the current algorithms address. These can take approaches from the multi-object tracking community, as in DeepSORT \cite{wojke_simple_2017}, where appearance uncertainty and localization and motion modelling are both taken into account for estimating the next location.  

The underwater-specific attributes, in \Cref{fig:env_attr_results}, however are more insightful. In the midwater and active lighting scenarios, which as previously noted are highly correlated in this dataset (that could be mitigated by collecting shallow night-time data), most trackers perform exceptionally well. This suggests that it may be reasonable to deploy semi-supervised trackers in deeper midwater environments that are actively lit, especially if the animals tend to be solitary. It is also interesting to note that the transformer-based architectures also perform dramatically better under those conditions, perhaps because of better appearance representations. 

By contrast, complex environments such as coral reefs and sea grass still pose exceptional challenges, as shown by \Cref{fig:env_attr_results} in the seabed (SB), coral reef (CR), and seagrass (SG) environments. The results are especially skewed downward in the case of seagrass. Across these attributes, online discriminators, especially KeepTrack and SuperDiMP, still are the most accurate.

In many cases, such as most of the fish species in CR environments, or the squid in rocky terrain, as can be seen in \Cref{fig:qual_results}, we found that all trackers were likely to fail when the animal was both in (A) \textit{specific orientations or body configurations (such as compressed body profiles in squid)} and (B) nearby complex backgrounds or similar objects. For example, most fish in the dataset are extremely thin when viewed from the back or top, in these perspectives they lack any distinguishing visual characteristics. The homogeneous colors in these settings, due to color absorption, and similar textures from the backgrounds, in the case of fish in reefs or squids in rocky terrain, tend to cause all the trackers to fail. This can be seen in the bluetang, angelfish, and squid examples in \Cref{fig:qual_results}.

One reason that the seagrass examples are difficult is that we have several octopus sequences, as shown in the fifth row of \Cref{fig:qual_results}, that were showing some amount of camouflage. 



Our experiments suggest that the grand challenge in the domain of marine animal tracking problems corresponds to tracking octopuses, both in fully-supervised or semi-supervised tracking contexts, because of their extreme adversarial nature to tracking. They have evolved stealth behaviors and unequalled rapid adaptive camouflage. On a typical forage, an octopus might change its camouflage more than 170 time/hour \cite{hanlon_crypsis_1999}. Moreover, the octopus can change its body shape dramatically with its 8 malleable arms, and it can also change its 3-D skin texture to match fine or course texture of adjacent corals, sponges, tunicates, algal epiphytes, etc. so that its ability to blend in is exceptional, even very experienced diving biologists are fooled. In our dataset however, the human annotators are still able to pick them out in the lower resolution dataset, so they present a clear target for future innovations.

We believe one major explanation for the success of online discriminative networks is to consider the problem from a perspective on long-term robustness vs. short-term accuracy. In the visual tracking problem, the best predictor of the appearance of an object at frame $t$ is its appearance at $t-1$. And so having a representation of this appearance model increases short-term accuracy. However, because only at time $t=0$ do we have a ground truth label, this is the most robust appearance model. This means that all decisions made on later object appearances are based on weak labels provided from the algorithm itself, this is prone to what is called concept drift in the online learning community \cite{mittal_online_nodate}. All the online discriminative-based trackers attempt to combat this issue by training not only on the most recent appearance, but also a history of appearances as well as augmentations of the initial image. However, many of these approaches are not well-principled. This trade-off is perhaps most easily seen in the shark track shown in \Cref{fig:qual_results}. Here, the Siamese-based networks are able to recover because they are robust to concept drift, whereas most trackers end up learning the appearance of the AUV that partially occludes the shark for a short time, even though the AUV was not specified in the original object appearance.

Though our dataset is relatively small, we hope to continue to contribute to it with more sequences in the future. We believe it provides a necessary and sufficient first step towards understanding performance of semi-supervised trackers in a variety of conditions specific to underwater domains. We can begin to distinguish between simpler and more difficult environments, species, and behaviors, so that AUV practitioners and the semi-supervised tracking community can begin to take more educated directions in their use in the wild. 

\subsection{Challenges in real-world marine animal tracking}

Overall, the real-world tracking results shown in \Cref{fig:barry}, \Cref{fig:jelly}, and \Cref{fig:jack} illustrate the exciting potential, but also some of the remaining challenges, of using semi-supervised trackers on vision-capable AUVs to perform longer-term marine animal tracking and monitoring, without the need for extensive labelled datasets. To the authors' knowledge, these demonstrations are among the first attempts to track complex marine animals with an AUV, in a difficult underwater setting, using only a visual semi-supervised tracker. There were \textit{no a-priori training of the targets or tagging}, and most of the tracking was \textit{autonomous}. The results from \Cref{fig:barry_trajectory} and \Cref{fig:jelly} especially highlight the benefits of applying these techniques to generate high-resolution data about animal trajectories in both novel environments and species, and even more so if stereo-vision can be incorporated in the near-term. For future tracks, it would be possible to generate reasonable 3D trajectories all in real-time. With stereo, marine biologists can then use these to estimate velocity and animal size, and in turn quantities such as energetics.

However, many challenges still exist before these can be used more reliably in other scenarios. We discuss some of our findings here based on our tracks of other animals, such as the jack in \Cref{fig:jack}, and even in some cases for \Cref{fig:barry}. While in some instances, while the jack was alone and swimming around the axis of the robot, as in Pane 1 in \Cref{fig:jack}, the tracker had no issues, though the robot was sometimes unable to rotate fast enough. In others, the tracker would either track another jack, as in Pane 2 and 3, because of their very similar appearance. Finally, the tracker would occasionally latch onto coral reefs, as in Pane 4 of \Cref{fig:jack} or the last frame of \Cref{fig:barry}, when the fish were rotated such that they faced away from the camera, their profile would become thin and resemble much of the fan corals around them. In these cases, the tracker would often, incorrectly, latch onto the corals.

In both the barracuda and jack tracks, we found many problems could be summarized by the following limitations:

\begin{itemize}
    \item \textbf{Limitations of the tracker}
    \begin{itemize}
        \item \textbf{Initialization issues} - importantly, as shown in the manual control plot of \Cref{fig:barry_trajectory}, though also present in the jack track as well, we often had to manually control the robot for the first several seconds of the track. This is because KeepTrackFast (and all the KeepTrack and DiMP-based trackers) have an initial training phase that can last several seconds on edge computing devices like the Jetson Xavier. In these cases, it is impossible to track the animal, and for smaller fish which rotate quickly, we often could not continue tracking after initialization. Additionally, initially selecting the bounding box is challenging. The human operator was using a mouse on a non-stationary and small boat and also experiencing latency over the tether, the human operator was unable to perfectly select the target while also maintaining sight of the animal unless it was mostly stationary to start. These issues are only experienced through real-world testing.
        \item \textbf{Track-time issues} - many of these issues were highlighted previously and discovered through our analysis on the VMAT dataset. Specifically, in cases where the tracked animals rotated or compressed into thin perspectives and were surrounded by either similar objects or complex backgrounds (such as other fish or corals), the trackers were unable to distinguish between the other objects and the target, and often failed in these cases. Extremely fast darting maneuvers, associated with both immediate changes in position and appearance, also tended to cause the tracker to fail. We believe that some investigation into improving speed on edge devices and more probabilistic localization techniques that rely less on appearance in these cases, as proposed in \Cref{sec:tracker_discussion}, would be helpful.
    \end{itemize}
    \item \textbf{Limitations of vehicle dynamics and other sensing} - the vehicle itself was often too slow to track some species of highly dynamic organisms such as smaller fish or sharks. In addition, many smaller fish spend most of their time close to, or in crevices of, the reef, which the vehicle is unable to approach for coral safety reasons. Many of these issues require hardware changes to the vehicle, or additional sensing techniques to better estimate distance from sensitive boundaries, such as the coral reefs, in order to better avoid them and navigate around them during tracking.
\end{itemize}

Our analysis of these algorithms on our VMAT dataset also helped both select an appropriate algorithm in terms of both \textit{accuracy} and \textit{speed}. It is important to note that KeepTrack was not useable on our AUV because it ran at only 2-4fps in our bench tests, which we believe is insufficient for real-world autonomous tracking, compared to 7-9fps of KeepTrackFast.

\section{Conclusion}\label{sec13}
\label{sec:conclusion}
Behavior of marine animals are inherently hard to study due to lack of availability of datasets characterizing their behavior in their natural environment. In this work we propose the use of a semi-supervised tracker on underwater robots to rapidly collect large datasets with minimal prior knowledge. Our contributions are on three fronts. First, we introduced a marine animal tracking dataset with 33 video sequences captured by mobile camera systems while following a marine animal. Second, we evaluate current state-of-the-art semi-supervised tracking algorithms over this dataset using novel evaluation metrics, specific to the underwater domain, which allow marine practitioners to better distinguish application-specific tradeoffs and capabilities of different trackers. Finally, we have, to the best of our knowledge, demonstrated the first use of a semi-supervised tracker onboard an AUV to track a  wide variety of marine animals (barracuda, jellyfish, jacks,..) spontaneously, in visually complex underwater environments. These demonstrations provides encouraging evidence towards the use of these technologies for marine animal tracking.

\bmhead{Supplementary information}
The full dataset is available at: \url{http://warp.whoi.edu/vmat/}. We also include the following supplemental videos for the AUV tracks:
\begin{itemize}
    \item \textbf{S1} \label{sm:1} barracuda tracking video
    \item \textbf{S2} \label{sm:2} jellyfish tracking video, 2022-10-31 at Tektite
    \item \textbf{S3} \label{sm:3} jellyfish tracking video, 2022-10-31 at Booby Rock
    \item \textbf{S4} \label{sm:4} jellyfish tracking video, 2022-11-1 at Booby Rock
    \item \textbf{S5} \label{sm:5} jack tracking video
\end{itemize}

\bmhead{Acknowledgments}
 Special thanks to Amy Kukulya and Roger Stokey at the Woods Hole Oceanographic Institution for the shark videos, the Mesobot team \cite{yoerger_hybrid_2021} for the solmissus and larvacean videos. Thanks to the WHOI WARPLab members Seth McCammon, John San Soucie, Stewart Jamieson, Daniel Yang, and Jessica E. Todd, Ethan Fahnestock for editing, and Cynthia Becker, Prajna Jandial, Nad\`ege Aoki, Nathan Formel, and Sierra Jarriel for species identification and support in the USVI data collection efforts. This work is supported in part by  The Investment in Science Fund at WHOI, and NSF NRI awards \# 1734400, 2133029. Levi Cai was supported by the NDSEG Fellowship. Also thanks to the NVIDIA Hardware Grant for a GPU for running evaluations.








\bibliography{sn-article}


\begin{thebibliography}{50}
\ifx \bisbn   \undefined \def \bisbn  #1{ISBN #1}\fi
\ifx \binits  \undefined \def \binits#1{#1}\fi
\ifx \bauthor  \undefined \def \bauthor#1{#1}\fi
\ifx \batitle  \undefined \def \batitle#1{#1}\fi
\ifx \bjtitle  \undefined \def \bjtitle#1{#1}\fi
\ifx \bvolume  \undefined \def \bvolume#1{\textbf{#1}}\fi
\ifx \byear  \undefined \def \byear#1{#1}\fi
\ifx \bissue  \undefined \def \bissue#1{#1}\fi
\ifx \bfpage  \undefined \def \bfpage#1{#1}\fi
\ifx \blpage  \undefined \def \blpage #1{#1}\fi
\ifx \burl  \undefined \def \burl#1{\textsf{#1}}\fi
\ifx \doiurl  \undefined \def \doiurl#1{\url{https://doi.org/#1}}\fi
\ifx \betal  \undefined \def \betal{\textit{et al.}}\fi
\ifx \binstitute  \undefined \def \binstitute#1{#1}\fi
\ifx \binstitutionaled  \undefined \def \binstitutionaled#1{#1}\fi
\ifx \bctitle  \undefined \def \bctitle#1{#1}\fi
\ifx \beditor  \undefined \def \beditor#1{#1}\fi
\ifx \bpublisher  \undefined \def \bpublisher#1{#1}\fi
\ifx \bbtitle  \undefined \def \bbtitle#1{#1}\fi
\ifx \bedition  \undefined \def \bedition#1{#1}\fi
\ifx \bseriesno  \undefined \def \bseriesno#1{#1}\fi
\ifx \blocation  \undefined \def \blocation#1{#1}\fi
\ifx \bsertitle  \undefined \def \bsertitle#1{#1}\fi
\ifx \bsnm \undefined \def \bsnm#1{#1}\fi
\ifx \bsuffix \undefined \def \bsuffix#1{#1}\fi
\ifx \bparticle \undefined \def \bparticle#1{#1}\fi
\ifx \barticle \undefined \def \barticle#1{#1}\fi
\bibcommenthead
\ifx \bconfdate \undefined \def \bconfdate #1{#1}\fi
\ifx \botherref \undefined \def \botherref #1{#1}\fi
\ifx \url \undefined \def \url#1{\textsf{#1}}\fi
\ifx \bchapter \undefined \def \bchapter#1{#1}\fi
\ifx \bbook \undefined \def \bbook#1{#1}\fi
\ifx \bcomment \undefined \def \bcomment#1{#1}\fi
\ifx \oauthor \undefined \def \oauthor#1{#1}\fi
\ifx \citeauthoryear \undefined \def \citeauthoryear#1{#1}\fi
\ifx \endbibitem  \undefined \def \endbibitem {}\fi
\ifx \bconflocation  \undefined \def \bconflocation#1{#1}\fi
\ifx \arxivurl  \undefined \def \arxivurl#1{\textsf{#1}}\fi
\csname PreBibitemsHook\endcsname

\bibitem{hanlon_crypsis_1999}
\begin{barticle}
\bauthor{\bsnm{Hanlon}, \binits{R.T.}},
\bauthor{\bsnm{Forsythe}, \binits{J.W.}},
\bauthor{\bsnm{Joneschild}, \binits{D.E.}}:
\batitle{Crypsis, conspicuousness, mimicry and polyphenism as antipredator
  defences of foraging octopuses on indo-pacific coral reefs, with a method of
  quantifying crypsis from video tapes}.
\bjtitle{Biological Journal of the Linnean Society}
\bvolume{66}(\bissue{1}),
\bfpage{1}--\blpage{22}
(\byear{1999}).
\doiurl{10.1006/bijl.1998.0264}.
Accessed 2022-04-29
\end{barticle}
\endbibitem

\bibitem{hanlon_flamboyant_2020}
\begin{barticle}
\bauthor{\bsnm{Hanlon}, \binits{R.T.}},
\bauthor{\bsnm{{McManus}}, \binits{G.}}:
\batitle{Flamboyant cuttlefish behavior: Camouflage tactics and complex
  colorful reproductive behavior assessed during field studies at lembeh
  strait, indonesia}.
\bjtitle{Journal of Experimental Marine Biology and Ecology}
\bvolume{529},
\bfpage{151397}
(\byear{2020}).
\doiurl{10.1016/j.jembe.2020.151397}.
Accessed 2022-04-29
\end{barticle}
\endbibitem

\bibitem{kukulya_3d_2015}
\begin{bchapter}
\bauthor{\bsnm{Kukulya}, \binits{A.L.}},
\bauthor{\bsnm{Stokey}, \binits{R.}},
\bauthor{\bsnm{Littlefield}, \binits{R.}},
\bauthor{\bsnm{Jaffre}, \binits{F.}},
\bauthor{\bsnm{Padilla}, \binits{E.M.H.}},
\bauthor{\bsnm{Skomal}, \binits{G.}}:
\bctitle{3d real-time tracking, following and imaging of white sharks with an
  autonomous underwater vehicle}.
In: \bbtitle{{OCEANS} 2015 - Genova},
pp. \bfpage{1}--\blpage{6}
(\byear{2015}).
\doiurl{10.1109/OCEANS-Genova.2015.7271546}
\end{bchapter}
\endbibitem

\bibitem{mooney_biologging_nodate}
\begin{bchapter}
\bauthor{\bsnm{Mooney}, \binits{T.A.}}:
\bctitle{Biologging ecology and oceanography: Integrative approaches to
  animal-bourne observations in a changing ocean}.
In: \bbtitle{Ocean Sciences Meeting 2020}
(\byear{2020})
\end{bchapter}
\endbibitem

\bibitem{priede_abyssal_2020}
\begin{barticle}
\bauthor{\bsnm{Priede}, \binits{I.G.}},
\bauthor{\bsnm{Drazen}, \binits{J.C.}},
\bauthor{\bsnm{Bailey}, \binits{D.M.}},
\bauthor{\bsnm{Kuhnz}, \binits{L.A.}},
\bauthor{\bsnm{Fabian}, \binits{D.}}:
\batitle{Abyssal demersal fishes recorded at station m (34 50n, 123 00w, 4100 m
  depth) in the northeast pacific ocean: An annotated check list and
  synthesis}.
\bjtitle{Deep Sea Research Part {II}: Topical Studies in Oceanography}
\bvolume{173},
\bfpage{104648}
(\byear{2020}).
\doiurl{10.1016/j.dsr2.2019.104648}.
Accessed 2022-04-29
\end{barticle}
\endbibitem

\bibitem{katija_visual_nodate-1}
\begin{botherref}
\oauthor{\bsnm{Katija}, \binits{K.}},
\oauthor{\bsnm{Roberts}, \binits{P.L.D.}},
\oauthor{\bsnm{Daniels}, \binits{J.}},
\oauthor{\bsnm{Lapides}, \binits{A.}},
\oauthor{\bsnm{Barnard}, \binits{K.}},
\oauthor{\bsnm{Risi}, \binits{M.}},
\oauthor{\bsnm{Ranaan}, \binits{B.Y.}},
\oauthor{\bsnm{Woodward}, \binits{B.G.}},
\oauthor{\bsnm{Takahashi}, \binits{J.}}:
Visual tracking of deepwater animals using machine learning-controlled robotic
  underwater vehicles.
IEEE WACV
(2021)
\end{botherref}
\endbibitem

\bibitem{bateson_2021}
\begin{botherref}
\oauthor{\bsnm{Bateson}, \binits{M.}},
\oauthor{\bsnm{Martin}, \binits{P.}}:
Measuring behaviour: An introductory guide
(2021)
\end{botherref}
\endbibitem

\bibitem{williams_surveying_2009}
\begin{bchapter}
\bauthor{\bsnm{Williams}, \binits{S.B.}},
\bauthor{\bsnm{Pizarro}, \binits{O.}},
\bauthor{\bsnm{How}, \binits{M.}},
\bauthor{\bsnm{Mercer}, \binits{D.}},
\bauthor{\bsnm{Powell}, \binits{G.}},
\bauthor{\bsnm{Marshall}, \binits{J.}},
\bauthor{\bsnm{Hanlon}, \binits{R.}}:
\bctitle{Surveying noctural cuttlefish camouflage behaviour using an {AUV}},
pp. \bfpage{214}--\blpage{219}
(\byear{2009}).
\doiurl{10.1109/ROBOT.2009.5152868}.
\bcomment{{ISSN}: 1050-4729}
\end{bchapter}
\endbibitem

\bibitem{yoerger_hybrid_2021}
\begin{barticle}
\bauthor{\bsnm{Yoerger}, \binits{D.R.}},
\bauthor{\bsnm{Govindarajan}, \binits{A.F.}},
\bauthor{\bsnm{Howland}, \binits{J.C.}},
\bauthor{\bsnm{Llopiz}, \binits{J.K.}},
\bauthor{\bsnm{Wiebe}, \binits{P.H.}},
\bauthor{\bsnm{Curran}, \binits{M.}},
\bauthor{\bsnm{Fujii}, \binits{J.}},
\bauthor{\bsnm{Gomez-Ibanez}, \binits{D.}},
\bauthor{\bsnm{Katija}, \binits{K.}},
\bauthor{\bsnm{Robison}, \binits{B.H.}},
\bauthor{\bsnm{Hobson}, \binits{B.W.}},
\bauthor{\bsnm{Risi}, \binits{M.}},
\bauthor{\bsnm{Rock}, \binits{S.M.}}:
\batitle{A hybrid underwater robot for multidisciplinary investigation of the
  ocean twilight zone}.
\bjtitle{AAAS Science Robotics}
(\byear{2021}).
\doiurl{10.1126/scirobotics.abe1901}.
\bcomment{Publisher: American Association for the Advancement of Science}
\end{barticle}
\endbibitem

\bibitem{fan_lasot_2020}
\begin{botherref}
\oauthor{\bsnm{Fan}, \binits{H.}},
\oauthor{\bsnm{Bai}, \binits{H.}},
\oauthor{\bsnm{Lin}, \binits{L.}},
\oauthor{\bsnm{Yang}, \binits{F.}},
\oauthor{\bsnm{Chu}, \binits{P.}},
\oauthor{\bsnm{Deng}, \binits{G.}},
\oauthor{\bsnm{Yu}, \binits{S.}},
\oauthor{\bsnm{Harshit}},
\oauthor{\bsnm{Huang}, \binits{M.}},
\oauthor{\bsnm{Liu}, \binits{J.}},
\oauthor{\bsnm{Xu}, \binits{Y.}},
\oauthor{\bsnm{Liao}, \binits{C.}},
\oauthor{\bsnm{Yuan}, \binits{L.}},
\oauthor{\bsnm{Ling}, \binits{H.}}:
{LaSOT}: A high-quality large-scale single object tracking benchmark
{\href{https://arxiv.org/abs/2009.03465}{{2009.03465}}}.
Accessed 2022-04-25
\end{botherref}
\endbibitem

\bibitem{akkaynak_sea-thru_nodate}
\begin{botherref}
\oauthor{\bsnm{Akkaynak}, \binits{D.}},
\oauthor{\bsnm{Treibitz}, \binits{T.}}:
Sea-thru: A method for removing water from underwater images.
IEEE CVPR
(2019)
\end{botherref}
\endbibitem

\bibitem{wang_underwater_2021}
\begin{barticle}
\bauthor{\bsnm{Wang}, \binits{Y.}},
\bauthor{\bsnm{Yu}, \binits{X.}},
\bauthor{\bsnm{An}, \binits{D.}},
\bauthor{\bsnm{Wei}, \binits{Y.}}:
\batitle{Underwater image enhancement and marine snow removal for fishery based
  on integrated dual-channel neural network}.
\bjtitle{Computers and Electronics in Agriculture}
\bvolume{186},
\bfpage{106182}
(\byear{2021}).
\doiurl{10.1016/j.compag.2021.106182}.
Accessed 2022-04-27
\end{barticle}
\endbibitem

\bibitem{kukulya_multi-vehicle_2016}
\begin{bchapter}
\bauthor{\bsnm{Kukulya}, \binits{A.L.}},
\bauthor{\bsnm{Stokey}, \binits{R.}},
\bauthor{\bsnm{Fiester}, \binits{C.}},
\bauthor{\bsnm{Padilla}, \binits{E.M.H.}},
\bauthor{\bsnm{Skomal}, \binits{G.}}:
\bctitle{Multi-vehicle autonomous tracking and filming of white sharks
  carcharodon carcharias}.
In: \bbtitle{2016 {IEEE}/{OES} Autonomous Underwater Vehicles ({AUV})},
pp. \bfpage{423}--\blpage{430}
(\byear{2016}).
\doiurl{10.1109/AUV.2016.7778707}.
\bcomment{{ISSN}: 2377-6536}
\end{bchapter}
\endbibitem

\bibitem{lin_focal_nodate}
\begin{botherref}
\oauthor{\bsnm{Lin}, \binits{T.-Y.}},
\oauthor{\bsnm{Goyal}, \binits{P.}},
\oauthor{\bsnm{Girshick}, \binits{R.}},
\oauthor{\bsnm{He}, \binits{K.}},
\oauthor{\bsnm{Dollar}, \binits{P.}}:
Focal loss for dense object detection.
IEEE ICCV
(2017)
\end{botherref}
\endbibitem

\bibitem{katija_fathomnet_2022}
\begin{botherref}
\oauthor{\bsnm{Katija}, \binits{K.}},
\oauthor{\bsnm{Orenstein}, \binits{E.}},
\oauthor{\bsnm{Schlining}, \binits{B.}},
\oauthor{\bsnm{Lundsten}, \binits{L.}},
\oauthor{\bsnm{Barnard}, \binits{K.}},
\oauthor{\bsnm{Sainz}, \binits{G.}},
\oauthor{\bsnm{Boulais}, \binits{O.}},
\oauthor{\bsnm{Cromwell}, \binits{M.}},
\oauthor{\bsnm{Butler}, \binits{E.}},
\oauthor{\bsnm{Woodward}, \binits{B.}},
\oauthor{\bsnm{Bell}, \binits{K.C.}}:
{FathomNet}: A global image database for enabling artificial intelligence in
  the ocean
(2022)
{\href{https://arxiv.org/abs/2109.14646}{{2109.14646}}}.
Accessed 2022-04-29
\end{botherref}
\endbibitem

\bibitem{dawkins_open-source_2017}
\begin{botherref}
\oauthor{\bsnm{Dawkins}, \binits{M.}},
\oauthor{\bsnm{Sherrill}, \binits{L.}},
\oauthor{\bsnm{Fieldhouse}, \binits{K.}},
\oauthor{\bsnm{Hoogs}, \binits{A.}},
\oauthor{\bsnm{Richards}, \binits{B.}},
\oauthor{\bsnm{Zhang}, \binits{D.}},
\oauthor{\bsnm{Prasad}, \binits{L.}},
\oauthor{\bsnm{Williams}, \binits{K.}},
\oauthor{\bsnm{Lauffenburger}, \binits{N.}},
\oauthor{\bsnm{Wang}, \binits{G.}}:
An open-source platform for underwater image and video analytics.
In: 2017 {IEEE} Winter Conference on Applications of Computer Vision ({WACV}),
pp. 898--906.
\doiurl{10.1109/WACV.2017.105}
\end{botherref}
\endbibitem

\bibitem{schlining_mbaris_2006}
\begin{botherref}
\oauthor{\bsnm{Schlining}, \binits{B.M.}},
\oauthor{\bsnm{Stout}, \binits{N.J.}}:
{MBARI}'s video annotation and reference system.
In: {OCEANS} 2006,
pp. 1--5.
\doiurl{10.1109/OCEANS.2006.306879}.
{ISSN}: 0197-7385
\end{botherref}
\endbibitem

\bibitem{noauthor_ozfish_nodate-1}
\begin{botherref}
{OzFish} Dataset - Machine Learning Dataset for Baited Remote Underwater Video
  Stations.
\url{https://apps.aims.gov.au/metadata/view/38c829d4-6b6d-44a1-9476-f9b0955ce0b8}
Accessed 2022-04-29
\end{botherref}
\endbibitem

\bibitem{saleh_realistic_2020-1}
\begin{botherref}
\oauthor{\bsnm{Saleh}, \binits{A.}},
\oauthor{\bsnm{Laradji}, \binits{I.H.}},
\oauthor{\bsnm{Konovalov}, \binits{D.A.}},
\oauthor{\bsnm{Bradley}, \binits{M.}},
\oauthor{\bsnm{Vazquez}, \binits{D.}},
\oauthor{\bsnm{Sheaves}, \binits{M.}}:
A realistic fish-habitat dataset to evaluate algorithms for underwater visual
  analysis.
Nature Scientific Reports
\textbf{10}(1),
14671.
\doiurl{10.1038/s41598-020-71639-x}.
Accessed 2022-04-29
\end{botherref}
\endbibitem

\bibitem{nam_learning_2016}
\begin{botherref}
\oauthor{\bsnm{Nam}, \binits{H.}},
\oauthor{\bsnm{Han}, \binits{B.}}:
Learning multi-domain convolutional neural networks for visual tracking.
In: 2016 {IEEE} Conference on Computer Vision and Pattern Recognition ({CVPR}),
pp. 4293--4302.
{IEEE}.
\doiurl{10.1109/CVPR.2016.465}.
\url{http://ieeexplore.ieee.org/document/7780834/}
Accessed 2022-04-29
\end{botherref}
\endbibitem

\bibitem{kristan_visual_2013}
\begin{bchapter}
\bauthor{\bsnm{Kristan}, \binits{M.}},
\bauthor{\bsnm{Pflugfelder}, \binits{R.}},
\bauthor{\bsnm{Leonardis}, \binits{A.}},
\bauthor{\bsnm{Matas}, \binits{J.}},
\bauthor{\bsnm{Porikli}, \binits{F.}},
\bauthor{\bsnm{Cehovin}, \binits{L.}},
\bauthor{\bsnm{Nebehay}, \binits{G.}},
\bauthor{\bsnm{Fernandez}, \binits{G.}},
\bauthor{\bsnm{Vojir}, \binits{T.}},
\bauthor{\bsnm{Gatt}, \binits{A.}},
\bauthor{\bsnm{Khajenezhad}, \binits{A.}},
\bauthor{\bsnm{Salahledin}, \binits{A.}},
\bauthor{\bsnm{Soltani-Farani}, \binits{A.}},
\bauthor{\bsnm{Zarezade}, \binits{A.}},
\bauthor{\bsnm{Petrosino}, \binits{A.}},
\bauthor{\bsnm{Milton}, \binits{A.}},
\bauthor{\bsnm{Bozorgtabar}, \binits{B.}},
\bauthor{\bsnm{Li}, \binits{B.}},
\bauthor{\bsnm{Chan}, \binits{C.S.}},
\bauthor{\bsnm{Heng}, \binits{C.}},
\bauthor{\bsnm{Ward}, \binits{D.}},
\bauthor{\bsnm{Kearney}, \binits{D.}},
\bauthor{\bsnm{Monekosso}, \binits{D.}},
\bauthor{\bsnm{Karaimer}, \binits{H.C.}},
\bauthor{\bsnm{Rabiee}, \binits{H.R.}},
\bauthor{\bsnm{Zhu}, \binits{J.}},
\bauthor{\bsnm{Gao}, \binits{J.}},
\bauthor{\bsnm{Xiao}, \binits{J.}},
\bauthor{\bsnm{Zhang}, \binits{J.}},
\bauthor{\bsnm{Xing}, \binits{J.}},
\bauthor{\bsnm{Huang}, \binits{K.}},
\bauthor{\bsnm{Lebeda}, \binits{K.}},
\bauthor{\bsnm{Cao}, \binits{L.}},
\bauthor{\bsnm{Maresca}, \binits{M.E.}},
\bauthor{\bsnm{Lim}, \binits{M.K.}},
\bauthor{\bsnm{El~Helw}, \binits{M.}},
\bauthor{\bsnm{Felsberg}, \binits{M.}},
\bauthor{\bsnm{Remagnino}, \binits{P.}},
\bauthor{\bsnm{Bowden}, \binits{R.}},
\bauthor{\bsnm{Goecke}, \binits{R.}},
\bauthor{\bsnm{Stolkin}, \binits{R.}},
\bauthor{\bsnm{Lim}, \binits{S.Y.}},
\bauthor{\bsnm{Maher}, \binits{S.}},
\bauthor{\bsnm{Poullot}, \binits{S.}},
\bauthor{\bsnm{Wong}, \binits{S.}},
\bauthor{\bsnm{Satoh}, \binits{S.}},
\bauthor{\bsnm{Chen}, \binits{W.}},
\bauthor{\bsnm{Hu}, \binits{W.}},
\bauthor{\bsnm{Zhang}, \binits{X.}},
\bauthor{\bsnm{Li}, \binits{Y.}},
\bauthor{\bsnm{Niu}, \binits{Z.}}:
\bctitle{The visual object tracking {VOT}2013 challenge results}.
In: \bbtitle{2013 {IEEE} International Conference on Computer Vision
  Workshops},
pp. \bfpage{98}--\blpage{111}
(\byear{2013}).
\doiurl{10.1109/ICCVW.2013.20}
\end{bchapter}
\endbibitem

\bibitem{tao_siamese_2016}
\begin{botherref}
\oauthor{\bsnm{Tao}, \binits{R.}},
\oauthor{\bsnm{Gavves}, \binits{E.}},
\oauthor{\bsnm{Smeulders}, \binits{A.W.M.}}:
Siamese instance search for tracking
(2016)
{\href{https://arxiv.org/abs/1605.05863}{{1605.05863}}}.
Accessed 2022-04-29
\end{botherref}
\endbibitem

\bibitem{li_siamrpn_2019}
\begin{botherref}
\oauthor{\bsnm{Li}, \binits{B.}},
\oauthor{\bsnm{Wu}, \binits{W.}},
\oauthor{\bsnm{Wang}, \binits{Q.}},
\oauthor{\bsnm{Zhang}, \binits{F.}},
\oauthor{\bsnm{Xing}, \binits{J.}},
\oauthor{\bsnm{Yan}, \binits{J.}}:
{SiamRPN}++: Evolution of siamese visual tracking with very deep networks.
In: 2019 {IEEE}/{CVF} Conference on Computer Vision and Pattern Recognition
  ({CVPR}),
pp. 4277--4286.
{IEEE}.
\doiurl{10.1109/CVPR.2019.00441}.
\url{https://ieeexplore.ieee.org/document/8954116/}
Accessed 2021-03-22
\end{botherref}
\endbibitem

\bibitem{wang_fast_2018}
\begin{botherref}
\oauthor{\bsnm{Wang}, \binits{Q.}},
\oauthor{\bsnm{Zhang}, \binits{L.}},
\oauthor{\bsnm{Bertinetto}, \binits{L.}},
\oauthor{\bsnm{Hu}, \binits{W.}},
\oauthor{\bsnm{Torr}, \binits{P.H.S.}}:
Fast online object tracking and segmentation: A unifying approach.
IEEE CVPR
(2019).
Accessed 2021-03-22
\end{botherref}
\endbibitem

\bibitem{danelljan_atom_2019}
\begin{botherref}
\oauthor{\bsnm{Danelljan}, \binits{M.}},
\oauthor{\bsnm{Bhat}, \binits{G.}},
\oauthor{\bsnm{Khan}, \binits{F.S.}},
\oauthor{\bsnm{Felsberg}, \binits{M.}}:
{ATOM}: Accurate tracking by overlap maximization
(2019)
{\href{https://arxiv.org/abs/1811.07628}{{1811.07628}}}.
Accessed 2021-02-20
\end{botherref}
\endbibitem

\bibitem{bhat_learning_2019}
\begin{botherref}
\oauthor{\bsnm{Bhat}, \binits{G.}},
\oauthor{\bsnm{Danelljan}, \binits{M.}},
\oauthor{\bsnm{Van~Gool}, \binits{L.}},
\oauthor{\bsnm{Timofte}, \binits{R.}}:
Learning discriminative model prediction for tracking.
In: 2019 {IEEE}/{CVF} International Conference on Computer Vision ({ICCV}),
pp. 6181--6190.
{IEEE}.
\doiurl{10.1109/ICCV.2019.00628}.
\url{https://ieeexplore.ieee.org/document/9010649/}
Accessed 2021-04-19
\end{botherref}
\endbibitem

\bibitem{danelljan_convolutional_2015}
\begin{botherref}
\oauthor{\bsnm{Danelljan}, \binits{M.}},
\oauthor{\bsnm{Hager}, \binits{G.}},
\oauthor{\bsnm{Khan}, \binits{F.S.}},
\oauthor{\bsnm{Felsberg}, \binits{M.}}:
Convolutional features for correlation filter based visual tracking.
In: 2015 {IEEE} International Conference on Computer Vision Workshop ({ICCVW}),
pp. 621--629.
{IEEE}.
\doiurl{10.1109/ICCVW.2015.84}.
\url{http://ieeexplore.ieee.org/document/7406433/}
Accessed 2019-07-22
\end{botherref}
\endbibitem

\bibitem{danelljan_eco_2017}
\begin{botherref}
\oauthor{\bsnm{Danelljan}, \binits{M.}},
\oauthor{\bsnm{Bhat}, \binits{G.}},
\oauthor{\bsnm{Khan}, \binits{F.S.}},
\oauthor{\bsnm{Felsberg}, \binits{M.}}:
{ECO}: Efficient convolution operators for tracking
(2017)
{\href{https://arxiv.org/abs/1611.09224}{{1611.09224}}}.
Accessed 2020-12-14
\end{botherref}
\endbibitem

\bibitem{chen_transformer_2021}
\begin{botherref}
\oauthor{\bsnm{Chen}, \binits{X.}},
\oauthor{\bsnm{Yan}, \binits{B.}},
\oauthor{\bsnm{Zhu}, \binits{J.}},
\oauthor{\bsnm{Wang}, \binits{D.}},
\oauthor{\bsnm{Yang}, \binits{X.}},
\oauthor{\bsnm{Lu}, \binits{H.}}:
Transformer tracking.
In: 2021 {IEEE}/{CVF} Conference on Computer Vision and Pattern Recognition
  ({CVPR}),
pp. 8122--8131.
{IEEE}.
\doiurl{10.1109/CVPR46437.2021.00803}.
\url{https://ieeexplore.ieee.org/document/9578609/}
Accessed 2022-04-25
\end{botherref}
\endbibitem

\bibitem{wang_transformer_2021}
\begin{botherref}
\oauthor{\bsnm{Wang}, \binits{N.}},
\oauthor{\bsnm{Zhou}, \binits{W.}},
\oauthor{\bsnm{Wang}, \binits{J.}},
\oauthor{\bsnm{Li}, \binits{H.}}:
Transformer meets tracker: Exploiting temporal context for robust visual
  tracking.
In: 2021 {IEEE}/{CVF} Conference on Computer Vision and Pattern Recognition
  ({CVPR}),
pp. 1571--1580.
{IEEE}.
\doiurl{10.1109/CVPR46437.2021.00162}.
\url{https://ieeexplore.ieee.org/document/9578157/}
Accessed 2022-04-25
\end{botherref}
\endbibitem

\bibitem{mueller_benchmark_2016-1}
\begin{botherref}
\oauthor{\bsnm{Mueller}, \binits{M.}},
\oauthor{\bsnm{Smith}, \binits{N.}},
\oauthor{\bsnm{Ghanem}, \binits{B.}}:
A benchmark and simulator for {UAV} tracking.
In: Leibe, B., Matas, J., Sebe, N., Welling, M. (eds.)
Computer Vision – {ECCV} 2016.
Lecture Notes in Computer Science,
pp. 445--461.
Springer.
\doiurl{10.1007/978-3-319-46448-0_27}
\end{botherref}
\endbibitem

\bibitem{huang_got-10k_2021}
\begin{barticle}
\bauthor{\bsnm{Huang}, \binits{L.}},
\bauthor{\bsnm{Zhao}, \binits{X.}},
\bauthor{\bsnm{Huang}, \binits{K.}}:
\batitle{{GOT}-10k: A large high-diversity benchmark for generic object
  tracking in the wild}.
\bjtitle{{IEEE} Transactions on Pattern Analysis and Machine Intelligence}
\bvolume{43}(\bissue{5}),
\bfpage{1562}--\blpage{1577}
(\byear{2021}).
\doiurl{10.1109/TPAMI.2019.2957464}
\end{barticle}
\endbibitem

\bibitem{kristan_eighth_2020}
\begin{botherref}
\oauthor{\bsnm{Kristan}, \binits{M.}},
\oauthor{\bsnm{Leonardis}, \binits{A.}},
\oauthor{\bsnm{Matas}, \binits{J.}},
\oauthor{\bsnm{Felsberg}, \binits{M.}},
\oauthor{\bsnm{Pflugfelder}, \binits{R.}},
\oauthor{\bsnm{Kamarainen}, \binits{J.-K.}},
\oauthor{\bsnm{Danelljan}, \binits{M.}},
\oauthor{\bsnm{Zajc}, \binits{L.C.}},
\oauthor{\bsnm{Lukezic}, \binits{A.}},
\oauthor{\bsnm{Drbohlav}, \binits{O.}},
\oauthor{\bsnm{He}, \binits{L.}},
\oauthor{\bsnm{Zhang}, \binits{Y.}},
\oauthor{\bsnm{Yan}, \binits{S.}},
\oauthor{\bsnm{Yang}, \binits{J.}},
\oauthor{\bsnm{Fernandez}, \binits{G.}},
\oauthor{\bsnm{Hauptmann}, \binits{A.}},
\oauthor{\bsnm{Memarmoghadam}, \binits{A.}},
\oauthor{\bsnm{Garcia-Martin}, \binits{A.}},
\oauthor{\bsnm{Robinson}, \binits{A.}},
\oauthor{\bsnm{Varfolomieiev}, \binits{A.}},
\oauthor{\bsnm{Gebrehiwot}, \binits{A.H.}},
\oauthor{\bsnm{Uzun}, \binits{B.}},
\oauthor{\bsnm{Yan}, \binits{B.}},
\oauthor{\bsnm{Li}, \binits{B.}},
\oauthor{\bsnm{Qian}, \binits{C.}},
\oauthor{\bsnm{Tsai}, \binits{C.-Y.}},
\oauthor{\bsnm{Micheloni}, \binits{C.}},
\oauthor{\bsnm{Wang}, \binits{D.}},
\oauthor{\bsnm{Wang}, \binits{F.}},
\oauthor{\bsnm{Xie}, \binits{F.}},
\oauthor{\bsnm{Lawin}, \binits{F.J.}},
\oauthor{\bsnm{Gustafsson}, \binits{F.}},
\oauthor{\bsnm{Foresti}, \binits{G.L.}},
\oauthor{\bsnm{Bhat}, \binits{G.}},
\oauthor{\bsnm{Chen}, \binits{G.}},
\oauthor{\bsnm{Ling}, \binits{H.}},
\oauthor{\bsnm{Zhang}, \binits{H.}},
\oauthor{\bsnm{Cevikalp}, \binits{H.}},
\oauthor{\bsnm{Zhao}, \binits{H.}},
\oauthor{\bsnm{Bai}, \binits{H.}},
\oauthor{\bsnm{Kuchibhotla}, \binits{H.C.}},
\oauthor{\bsnm{Saribas}, \binits{H.}},
\oauthor{\bsnm{Fan}, \binits{H.}},
\oauthor{\bsnm{Ghanei-Yakhdan}, \binits{H.}},
\oauthor{\bsnm{Li}, \binits{H.}},
\oauthor{\bsnm{Peng}, \binits{H.}},
\oauthor{\bsnm{Lu}, \binits{H.}},
\oauthor{\bsnm{Li}, \binits{H.}},
\oauthor{\bsnm{Khaghani}, \binits{J.}},
\oauthor{\bsnm{Bescos}, \binits{J.}},
\oauthor{\bsnm{Li}, \binits{J.}},
\oauthor{\bsnm{Fu}, \binits{J.}},
\oauthor{\bsnm{Yu}, \binits{J.}},
\oauthor{\bsnm{Xu}, \binits{J.}},
\oauthor{\bsnm{Kittler}, \binits{J.}},
\oauthor{\bsnm{Yin}, \binits{J.}},
\oauthor{\bsnm{Lee}, \binits{J.}},
\oauthor{\bsnm{Yu}, \binits{K.}},
\oauthor{\bsnm{Liu}, \binits{K.}},
\oauthor{\bsnm{Yang}, \binits{K.}},
\oauthor{\bsnm{Dai}, \binits{K.}},
\oauthor{\bsnm{Cheng}, \binits{L.}},
\oauthor{\bsnm{Zhang}, \binits{L.}},
\oauthor{\bsnm{Wang}, \binits{L.}},
\oauthor{\bsnm{Wang}, \binits{L.}},
\oauthor{\bsnm{Van~Gool}, \binits{L.}},
\oauthor{\bsnm{Bertinetto}, \binits{L.}},
\oauthor{\bsnm{Dunnhofer}, \binits{M.}},
\oauthor{\bsnm{Cheng}, \binits{M.}},
\oauthor{\bsnm{Dasari}, \binits{M.M.}},
\oauthor{\bsnm{Wang}, \binits{N.}},
\oauthor{\bsnm{Wang}, \binits{N.}},
\oauthor{\bsnm{Zhang}, \binits{P.}},
\oauthor{\bsnm{Torr}, \binits{P.H.S.}},
\oauthor{\bsnm{Wang}, \binits{Q.}},
\oauthor{\bsnm{Timofte}, \binits{R.}},
\oauthor{\bsnm{Gorthi}, \binits{R.K.S.}},
\oauthor{\bsnm{Choi}, \binits{S.}},
\oauthor{\bsnm{Marvasti-Zadeh}, \binits{S.M.}},
\oauthor{\bsnm{Zhao}, \binits{S.}},
\oauthor{\bsnm{Kasaei}, \binits{S.}},
\oauthor{\bsnm{Qiu}, \binits{S.}},
\oauthor{\bsnm{Chen}, \binits{S.}},
\oauthor{\bsnm{Schön}, \binits{T.B.}},
\oauthor{\bsnm{Xu}, \binits{T.}},
\oauthor{\bsnm{Lu}, \binits{W.}},
\oauthor{\bsnm{Hu}, \binits{W.}},
\oauthor{\bsnm{Zhou}, \binits{W.}},
\oauthor{\bsnm{Qiu}, \binits{X.}},
\oauthor{\bsnm{Ke}, \binits{X.}},
\oauthor{\bsnm{Wu}, \binits{X.-J.}},
\oauthor{\bsnm{Zhang}, \binits{X.}},
\oauthor{\bsnm{Yang}, \binits{X.}},
\oauthor{\bsnm{Zhu}, \binits{X.}},
\oauthor{\bsnm{Jiang}, \binits{Y.}},
\oauthor{\bsnm{Wang}, \binits{Y.}},
\oauthor{\bsnm{Chen}, \binits{Y.}},
\oauthor{\bsnm{Ye}, \binits{Y.}},
\oauthor{\bsnm{Li}, \binits{Y.}},
\oauthor{\bsnm{Yao}, \binits{Y.}},
\oauthor{\bsnm{Lee}, \binits{Y.}},
\oauthor{\bsnm{Gu}, \binits{Y.}},
\oauthor{\bsnm{Wang}, \binits{Z.}},
\oauthor{\bsnm{Tang}, \binits{Z.}},
\oauthor{\bsnm{Feng}, \binits{Z.-H.}},
\oauthor{\bsnm{Mai}, \binits{Z.}},
\oauthor{\bsnm{Zhang}, \binits{Z.}},
\oauthor{\bsnm{Wu}, \binits{Z.}},
\oauthor{\bsnm{Ma}, \binits{Z.}}:
The eighth visual object tracking {VOT}2020 challenge results.
In: Bartoli, A., Fusiello, A. (eds.)
Computer Vision – {ECCV} 2020 Workshops,
pp. 547--601.
Springer.
\doiurl{10.1007/978-3-030-68238-5_39}
\end{botherref}
\endbibitem

\bibitem{wu_object_2015}
\begin{barticle}
\bauthor{\bsnm{Wu}, \binits{Y.}},
\bauthor{\bsnm{Lim}, \binits{J.}},
\bauthor{\bsnm{Yang}, \binits{M.-H.}}:
\batitle{Object tracking benchmark}.
\bjtitle{{IEEE} Transactions on Pattern Analysis and Machine Intelligence}
\bvolume{37}(\bissue{9}),
\bfpage{1834}--\blpage{1848}
(\byear{2015}).
\doiurl{10.1109/TPAMI.2014.2388226}
\end{barticle}
\endbibitem

\bibitem{caelles_2019_2019}
\begin{botherref}
\oauthor{\bsnm{Caelles}, \binits{S.}},
\oauthor{\bsnm{Pont-Tuset}, \binits{J.}},
\oauthor{\bsnm{Perazzi}, \binits{F.}},
\oauthor{\bsnm{Montes}, \binits{A.}},
\oauthor{\bsnm{Maninis}, \binits{K.-K.}},
\oauthor{\bsnm{Van~Gool}, \binits{L.}}:
The 2019 {DAVIS} challenge on {VOS}: Unsupervised multi-object segmentation
(2019)
{\href{https://arxiv.org/abs/1905.00737}{{1905.00737}}}.
Accessed 2021-03-23
\end{botherref}
\endbibitem

\bibitem{xu_youtube-vos_2018-1}
\begin{botherref}
\oauthor{\bsnm{Xu}, \binits{N.}},
\oauthor{\bsnm{Yang}, \binits{L.}},
\oauthor{\bsnm{Fan}, \binits{Y.}},
\oauthor{\bsnm{Yue}, \binits{D.}},
\oauthor{\bsnm{Liang}, \binits{Y.}},
\oauthor{\bsnm{Yang}, \binits{J.}},
\oauthor{\bsnm{Huang}, \binits{T.}}:
{YouTube}-{VOS}: A large-scale video object segmentation benchmark
{\href{https://arxiv.org/abs/1809.03327}{{1809.03327}}}.
Accessed 2022-04-29
\end{botherref}
\endbibitem

\bibitem{galoogahi_need_2017}
\begin{botherref}
\oauthor{\bsnm{Galoogahi}, \binits{H.K.}},
\oauthor{\bsnm{Fagg}, \binits{A.}},
\oauthor{\bsnm{Huang}, \binits{C.}},
\oauthor{\bsnm{Ramanan}, \binits{D.}},
\oauthor{\bsnm{Lucey}, \binits{S.}}:
Need for speed: A benchmark for higher frame rate object tracking.
In: 2017 {IEEE} International Conference on Computer Vision ({ICCV}),
pp. 1134--1143.
{IEEE}.
\doiurl{10.1109/ICCV.2017.128}.
\url{http://ieeexplore.ieee.org/document/8237390/}
Accessed 2022-04-29
\end{botherref}
\endbibitem

\bibitem{valmadre_long-term_2018}
\begin{botherref}
\oauthor{\bsnm{Valmadre}, \binits{J.}},
\oauthor{\bsnm{Bertinetto}, \binits{L.}},
\oauthor{\bsnm{Henriques}, \binits{J.F.}},
\oauthor{\bsnm{Tao}, \binits{R.}},
\oauthor{\bsnm{Vedaldi}, \binits{A.}},
\oauthor{\bsnm{Smeulders}, \binits{A.}},
\oauthor{\bsnm{Torr}, \binits{P.}},
\oauthor{\bsnm{Gavves}, \binits{E.}}:
Long-term tracking in the wild: A benchmark.
IEEE ECCV
(2018).
Accessed 2021-03-22
\end{botherref}
\endbibitem

\bibitem{noauthor_labelbox_nodate}
\begin{botherref}
Labelbox: The Leading Training Data Platform for Data Labeling.
\url{https://labelbox.com/}
Accessed 2022-04-29
\end{botherref}
\endbibitem

\bibitem{zhu_distractor-aware_2018}
\begin{botherref}
\oauthor{\bsnm{Zhu}, \binits{Z.}},
\oauthor{\bsnm{Wang}, \binits{Q.}},
\oauthor{\bsnm{Li}, \binits{B.}},
\oauthor{\bsnm{Wu}, \binits{W.}},
\oauthor{\bsnm{Yan}, \binits{J.}},
\oauthor{\bsnm{Hu}, \binits{W.}}:
Distractor-aware siamese networks for visual object tracking
(2018)
{\href{https://arxiv.org/abs/1808.06048}{{1808.06048}}}.
Accessed 2021-03-15
\end{botherref}
\endbibitem

\bibitem{danelljan_probabilistic_2020}
\begin{botherref}
\oauthor{\bsnm{Danelljan}, \binits{M.}},
\oauthor{\bsnm{Gool}, \binits{L.V.}},
\oauthor{\bsnm{Timofte}, \binits{R.}}:
Probabilistic regression for visual tracking.
In: 2020 {IEEE}/{CVF} Conference on Computer Vision and Pattern Recognition
  ({CVPR}),
pp. 7181--7190.
\doiurl{10.1109/CVPR42600.2020.00721}.
{ISSN}: 2575-7075
\end{botherref}
\endbibitem

\bibitem{mayer_learning_2021}
\begin{botherref}
\oauthor{\bsnm{Mayer}, \binits{C.}},
\oauthor{\bsnm{Danelljan}, \binits{M.}},
\oauthor{\bsnm{Pani~Paudel}, \binits{D.}},
\oauthor{\bsnm{Van~Gool}, \binits{L.}}:
Learning target candidate association to keep track of what not to track.
In: 2021 {IEEE}/{CVF} International Conference on Computer Vision ({ICCV}),
pp. 13424--13434.
{IEEE}.
\doiurl{10.1109/ICCV48922.2021.01319}.
\url{https://ieeexplore.ieee.org/document/9710884/}
Accessed 2022-04-29
\end{botherref}
\endbibitem

\bibitem{he_deep_2016}
\begin{botherref}
\oauthor{\bsnm{He}, \binits{K.}},
\oauthor{\bsnm{Zhang}, \binits{X.}},
\oauthor{\bsnm{Ren}, \binits{S.}},
\oauthor{\bsnm{Sun}, \binits{J.}}:
Deep residual learning for image recognition.
In: 2016 {IEEE} Conference on Computer Vision and Pattern Recognition ({CVPR}),
pp. 770--778.
\doiurl{10.1109/CVPR.2016.90}.
\url{http://ieeexplore.ieee.org/document/7780459/}
Accessed 2022-04-29
\end{botherref}
\endbibitem

\bibitem{chatfield_return_2014}
\begin{botherref}
\oauthor{\bsnm{Chatfield}, \binits{K.}},
\oauthor{\bsnm{Simonyan}, \binits{K.}},
\oauthor{\bsnm{Vedaldi}, \binits{A.}},
\oauthor{\bsnm{Zisserman}, \binits{A.}}:
Return of the devil in the details: Delving deep into convolutional nets.
British Machine Vision Conference (BMVC)
(2014).
Accessed 2022-04-29
\end{botherref}
\endbibitem

\bibitem{vaswani_attention_2017}
\begin{botherref}
\oauthor{\bsnm{Vaswani}, \binits{A.}},
\oauthor{\bsnm{Shazeer}, \binits{N.}},
\oauthor{\bsnm{Parmar}, \binits{N.}},
\oauthor{\bsnm{Uszkoreit}, \binits{J.}},
\oauthor{\bsnm{Jones}, \binits{L.}},
\oauthor{\bsnm{Gomez}, \binits{A.N.}},
\oauthor{\bsnm{Kaiser}, \binits{L.}},
\oauthor{\bsnm{Polosukhin}, \binits{I.}}:
Attention is all you need.
In: Advances in Neural Information Processing Systems,
vol. 30.
Curran Associates, Inc.
\url{https://proceedings.neurips.cc/paper/2017/hash/3f5ee243547dee91fbd053c1c4a845aa-Abstract.html}
Accessed 2022-04-29
\end{botherref}
\endbibitem

\bibitem{ferrari_trackingnet_2018-1}
\begin{bchapter}
\bauthor{\bsnm{Müller}, \binits{M.}},
\bauthor{\bsnm{Bibi}, \binits{A.}},
\bauthor{\bsnm{Giancola}, \binits{S.}},
\bauthor{\bsnm{Alsubaihi}, \binits{S.}},
\bauthor{\bsnm{Ghanem}, \binits{B.}}:
\bctitle{{TrackingNet}: A large-scale dataset and benchmark for object tracking
  in the wild}.
In: \bbtitle{European Conference on Computer Vision (ECCV)}
vol. \bseriesno{11205},
pp. \bfpage{310}--\blpage{327}
(\byear{2018}).
\doiurl{10.1007/978-3-030-01246-5_19}.
\burl{http://link.springer.com/10.1007/978-3-030-01246-5_19}
Accessed 2021-04-23
\end{bchapter}
\endbibitem

\bibitem{girdhar_icra_2023}
\begin{bchapter}
\bauthor{\bsnm{Girdhar}, \binits{Y.}},
\bauthor{\bsnm{McGuire}, \binits{N.}},
\bauthor{\bsnm{Cai}, \binits{L.}},
\bauthor{\bsnm{Jamieson}, \binits{S.}},
\bauthor{\bsnm{McCammon}, \binits{S.}},
\bauthor{\bsnm{Claus}, \binits{B.}},
\bauthor{\bsnm{Soucie}, \binits{J.E.S.}},
\bauthor{\bsnm{Todd}, \binits{J.E.}},
\bauthor{\bsnm{Mooney}, \binits{T.A.}}:
\bctitle{{CUREE}: A curious underwater robot for ecosystem exploration}.
In: \bbtitle{{IEEE} International Conference on Robotics and Automation
  ({ICRA})}
(\byear{2023}).
\bcomment{[To appear]}
\end{bchapter}
\endbibitem

\bibitem{noauthor_ros_nodate}
\begin{botherref}
{ROS}: Home.
\url{https://www.ros.org/}
Accessed 2022-04-29
\end{botherref}
\endbibitem

\bibitem{wojke_simple_2017}
\begin{botherref}
\oauthor{\bsnm{Wojke}, \binits{N.}},
\oauthor{\bsnm{Bewley}, \binits{A.}},
\oauthor{\bsnm{Paulus}, \binits{D.}}:
Simple online and realtime tracking with a deep association metric
(2017)
{\href{https://arxiv.org/abs/1703.07402}{{1703.07402}}}.
Accessed 2021-01-09
\end{botherref}
\endbibitem

\bibitem{mittal_online_nodate}
\begin{barticle}
\bauthor{\bsnm{Mittal}, \binits{V.}},
\bauthor{\bsnm{Kashyap}, \binits{I.}}:
\batitle{Online methods of learning in occurrence of concept drift}.
\bjtitle{International Journal of Computer Applications}
\bvolume{117}(\bissue{13}),
\bfpage{18}--\blpage{22}
(\byear{2015})
\end{barticle}
\endbibitem

\end{thebibliography}


\end{document}